\documentclass[twoside,11pt]{article}
\usepackage{jair,rawfonts}
\usepackage{amsmath}
\usepackage{amssymb}
\usepackage{stmaryrd}
\usepackage{times}
\usepackage{graphicx}

\usepackage{array}
\newtheorem{definition}{Definition}
\begin{document}
\title{Social Practices: a Complete Formalization}
\author{\name F.P.M. Dignum \email dignum@cs.umu.se\\
\addr Ume\aa\ University \&  Utrecht University \\
 Ume\aa, Sweden \& Utrecht, The Netherlands}
\maketitle

\begin{abstract}
Multi-agent models are a suitable starting point to model complex social interactions. However, as the complexity of the systems increase, we argue that novel modeling approaches are needed that can deal with inter-dependencies at different levels of society, where many heterogeneous parties (software agents, robots, humans) are interacting and reacting to each other. In this paper, we present a formalization of a social framework for agents based on the concept of \emph{Social Practices} as high level specifications of normal (expected) behavior in a given social context. We argue that social practices facilitate the practical reasoning of agents in standard social interactions. Thus they can support deliberations for complex situations just like conventions and norms. However, they also come with a social context that gives handles for social planning and deliberation in top of the normal functional deliberation. The main goal of this paper is to give a formalization of social practices that can be used as a basis for implementations and defining precise structures within which social learning can take place.
\end{abstract}

\section{Introduction and Motivation}

Interactions do not exist in a vacuum, but are surrounded by many social and physical constructs that shape and constrain that interaction \cite{Argyle}. Understanding and modeling the context of interactions is essential in the design of systems that are both realistic and computationally feasible. In fact, if context is not properly considered  \emph{``the patterns and the outcomes of the interactions are inherently unpredictable, and predicting the behavior of the overall system based on its constituent components is extremely difficult (sometimes impossible) because of the high likelihood of emergent (and unwanted) behavior"}~\cite{jennings:00}. E.g. imagine the situation that normally Fred takes the children to school on the way to work and his wife Sonja picks up her colleague and goes to work in a different direction. On a day that Fred has an important presentation to give in the morning, Sonja might decide to take the children to school in order to give Fred more time and peace of mind for his presentation. Even though it might not be the most cost effective solution, given the social context, it is the best overall behavior. It also has effect on the relation between Fred and Sonja and future interactions.
In human societies, we use social practices as means to cope with uncertainty of outcome of interaction. Interactions are embedded in a broad network of societal and institutional contexts, and social practices provide means to deal with this complexity. Endowing agents with means to represent and reason with social practices will enable smooth, flexible, context-aware human-agent interaction.
% * <jvazquez@cs.upc.edu> 2017-02-18T06:14:04.520Z:
% 
% I checked the use of "supple" and it seems that it is used for objects or materials, not for interactions. I could not find in google any use of "supple interaction".
% 
% ^ <m.v.dignum@tudelft.nl> 2017-02-18T14:14:10.218Z.

Applied to human-agent interactions, such as companion robots, human-agent-robot teamwork, or persuasive applications, social practices can simplify and guide the deliberation of the agent in complex contexts. In these domains, the agent is expected to support the user but there are no predefined protocols or fixed objectives, so agent actions must be based on a close observation of the situation and evaluation of possible user aims, and reaction to user actions and environment changes. This is exactly the reason for social practices in human interactions. Social practices refer to everyday practices and the way these are typically  performed by (most of) the members in the society. Although conventions and norms have similar aims as social practices, we will show that for the kind of applications where the human-agent interaction does not consist of a single action or action sequence, social practices provide an added value. Even though the subject of emergence and evolution of social practices is an important one, there is very little formal theory available on this aspect and thus for simplicity in this paper we assume social practices to be given and fixed. I.e. we will not elaborate on how the practice come to be, or how they are maintained and shared. Our conceptual agent architecture therefore takes a set of social practices as given, and the focus is on how the agent uses practices in its deliberation and planning. However, in the discussion section we will sketch some possible ways in which the framework for social practices as we will develop it in this paper can be made more dynamic.

In the core of this paper, we present a formalization of the theory of social practices as proposed in \cite{shove2012,Reckwitz} to be used as basis for a cognitive architecture. Our approach can be seen as a middle way between  using fixed interaction scripts and full free deliberation reactive to user and environment sensing. We build on the concept of landmark \cite{Dignum2007} in the sense, that social practices enable the specification of possible (abstract) plans without needing to represent every single component of the interaction. An agent will adopt a social practice, based on its evaluation of the context, and \textit{`fill-in the gaps'}, using its own capabilities and according to its own intentions. We provide a logic semantics for social practices, based on modalities to represent expectation, capability, roles, norms and contexts.

The objective of this paper is to provide a first step towards an architecture for social intelligent agents that use social contexts as the main starting point for interaction. In particular, we describe a formal specification of the constructs needed to represent and reason with social practices,  such that it fulfils the objectives of the agent, without needing to represent every possible aspect of the interaction, or path in a protocol.

In the next section we will give some more background on social practice theory and also will discuss how social practices relate to other social concepts such as conventions and norms. In section \ref{characteristics} we describe the main characteristics of social practices, translating from the rather vague notions used in social science to more concrete concepts more akin to computer science applications. In section \ref{formal} we develop the formal logic necessary to formalize all aspects of social practices as given in section \ref{characteristics}.  Then we can start to give a formal account of all the elements of social practices in section \ref{socialpracticeformal}. In section \ref{applications} we show how this formalization can be used by an agent to deliberate about a social practice and use it for guiding its behavior. In section \ref{conclusion} we discuss the use of the framework as a structure that can be used as a basis for emerging and maintaining social practices and we finish with some conclusions and future work.

\section{Background}
\label{background}
Social practice theory comes forth from a variety of different disciplines. It started from philosophical sociology with proponents like Bourdieu \cite{Bourdieu1976} and Giddens \cite{Giddens1979}. Later on Reckwitz and Shove \cite{Reckwitz,shove2012} have expanded on these ideas and also Schatzki \cite{Schatzki2012} made some valuable contributions. An important claim that all these people make is that important features of human life should be understood in terms of organized constellations of interacting persons. Thus people are not just creating these practices, but our deliberations are also based on the fact that most of our life is shaped by social practices. Thus we use social practices to categorize situations and decide upon ways of behaviour based on social practices. The main intuition behind this is that our life is quite cyclic in that many activities come back with a certain regularity. We have meals every day, go to work on Monday until Friday, go to the supermarket once a week, etc. These so-called Patterns of Life (\cite{Flosom2014}) can be exploited to create standard situations and expectations. It makes sense to categorize recurrent situations as social practices with a kind of standard behaviour for each of them.
We will discuss some background and related work from both social science on social practices and in the subsequent section from computer science on structures that seem related or might be used to represent social practices.
\subsection{Social Practice Theory}
Social practices are defined as accepted ways of doing things, contextual and materially mediated, that are shared between actors and routinized over time \cite{Reckwitz}. They can be seen as patterns which can be filled in by a multitude of single and often unique actions. Through (joint) performance, the patterns provided by the practice are filled out and reproduced.

According to \cite{Reckwitz,shove2012} a social practice consists of three parts: 
\begin{itemize}
\item Material: covers all physical aspects of the performance of a practice, including the human body (relates to physical aspects of a situation).
\item Meaning: refers to the issues which are considered to be relevant with respect to that material, i.e. understandings, beliefs and emotions (relates to social aspects of a situation).
\item Competence:  refers to skills and knowledge which are required to perform the practice (relates to the notion of deliberation about a situation).
\end{itemize}

The material for taking the children to school would be the route between home and school, the car used (is it a big or small one), the time being in the morning, the parents and children,the school bags, etc.\\ 
The competences denote kinds of activities that can be expected from the participants. They do not dictate a complete workflow or script for the social practice, but rather indicate expected actions at points in time. E.g. the children have to put on their coats and get their school bags before getting in the car. But putting on coats might not be needed in summer time. The individuals participating in the social practice are free to act in any way that conforms to the practice, and even to violate it if they wish. So, taking the children to school means that the children have to get in the car before driving to school. The exact route to school might differ depending on traffic. Also the conversations in the car are completely ``free". The activities also encompass the expected competencies of the actors involved. E.g. we expect that the parents know how to drive a car.\\
The meaning of a social practice couples the interaction perspective of it with the social perspective of the practice. E.g. letting the oldest child sit in front or the one who has her birthday indicates the special position and responsibilities of that child. Singing some songs with the children in the car might also help bonding with the children and facilitate later chores.

From the above description it can already be seen that social practices are more encompassing than conventions and norms. Conventions focus on the strategic advantage that an individual gets by conforming to the convention. The reason to follow a convention is that if all parties involved comply a kind of optimal coordination is reached. I.e. if we all drive on the left side of the road traffic will be more smooth than when everyone chooses the side to drive on freely. Thus conventions focus on the actual actions being performed and how they optimize the coordination. Social practices do not necessarily optimize the coordination. Because they indicate a type of expected actions and interactions given a social and physical context they will smoothen the coordination. However, this is not necessarily the optimal way the coordination could have been done. E.g. if we go to a presentation we sit down as soon as we see chairs standing in rows in the room. However, we could also keep standing (as is often done outside).

Social practices are different from norms. Norms are also applicable in certain situations and for particular people (or roles) and they also create expectations (nl. that the norm is followed). However, norms usually dictate a very specific behaviour rather than creating a set of loosely couple expectations as is the case for social practices. E.g. if the norm states that a car has to stop for a red light, it gives a very specific directive. If a norm is more abstract (like ``drive carefully") then we need to translate this into concrete norms for specific situations. This is different from saying that some parts of a situation are governed by the norm and others are still free of the normative influence. Basically, when specifying the norm one indicates exactly when the norm is applicable rather than general situations for which the norm can be applied in some part.

The resources describe the physical and social setting in which the practice takes place. Therefore it is the cognitive trigger of the practice and can also be used to check which objects and people are available to peform the expected actions within the social practice. Likewise the activities indicate which actions and action patterns are available for the planning of the individuals participating in the social practice. 
The meaning of a practice is related to the social perspective as it indicates the social affect and purpose of a social practice given that it is performed in a certain way. E.g. taking the children to school in a big SUV might signal the wealth or status (or stupidity) of the parents. Using the bike can signal the environmental consciousness of the parents. Also actions within a practice can get a specific social interpretation. E.g. asking a question during a lecture on a difficult topic can indicate interest and showing effort to learn. This might lead to more support from a lecturer later during the course.
Given the above descriptions of social practices one can summarize the purpose of them for individuals (informally) as follows:
\[ 
Perceive(resources) \Rightarrow Expect(activities) \wedge Expect(competencies)
\]
\[
Done(activities) \Rightarrow physical(postconditions) \wedge social(postconditions)
\]
Given a certain situation that is perceived in which the social practice is activated, the social practice now triggers expectations that the activities of that social practice will be executed. This implies that we assume a certain competence of all the people involved in the social practice (e.g. a parent can drive a car). After the activities have been executed, we do not just assume the postconditions of the actions hold, but also assume that certain social effects have been achieved according to the ``meaning" of the practice. Thus the social practice allows us to make a whole set of assumptions and have expectations that could otherwise not be readily made or would take a lot of effort to derive.

The dynamic nature of social practices is emphasized by Shove \cite{shove2012}: Each time it is used, elements of the practice, including know-how, meanings and purposes, are reconfigured and adapted. Therefore the use of social practices includes a constant learning of the individuals using the social practice in ever changing contexts. In this way social practices guide the learning process of agents in a natural way. In \cite{shove2012} the social aspect of social practices is emphasized by giving the social practice center stage in interactions and letting individuals be supporters of the social practice. It shows that social practices are shared (social) concepts that also exist outside individuals. The mere fact that they are shared and jointly created and maintained means that individuals playing a role in a social practice will expect certain behavior and reactions of the other participants in the social practice. Thus it is this aspect that makes the social practices so suitable for use in individual deliberation in social situations. In section \ref{applications} we will show how this can be done in our framework.

\subsection{Related Work from Computer Science}
In this section, we discuss several existing frameworks in computer science that seem related to the concept of social practices and show why they are not suitable or too limited to be used directly.\\
One framework that seems very close to social practices is the notion of \emph{scripts}.However, social practices are not just mere scripts in the sense of \cite{Minsky}. Practices are more flexible than the classical frames defined by scripts in that they can be extended and changed by learning and the "slots" only need to be filled in as far as they are needed to determine a course of action. Using these structures changes planning in many common situations to pattern recognition and filling in parameters. They support, rather than restrict deliberation about behaviour. E.g. the social practice of ``going to work'' incorporates usual means of transport that can be used, timing constraints, weather and traffic conditions, etc. So, normally you take a car to work, but if the weather is exceptionally bad the social practice does not force the default action, but rather gives input for deliberation about a new plan in this situation and take a bus or train (or even stay home). So, social practices can be seen as a kind of flexible script. Moreover, scripts do not incorporate any social meaning for the activities performed in them as social practices do provide.

\subsubsection{Agent Programming}
Multi-agent programming languages such as 2APL support the implementation of individual agents that can perform high-level reasoning and deliberation about their information (i.e., beliefs) and objectives (i.e., goals to achieve) in order to decide what actions to perform~\cite{dastani20082apl}. In order to reach its goals, an agent adopts
plans. 2APL provides programming constructs to implement beliefs, goals, actions, plans, events, and three different types of rules that can be applied to generate plans. In particular, 2APL  provides \textit{planning goal rules} that implement practical reasoning rules that can be used to generate plans for achieving goals and \textit{practical reasoning rules}, which can be used to expand abstract plans to concrete sequences of actions and to rewrite plans to cope with unforeseen circumstances~\cite{dastani2aplmodular09}.

A possible way to represent the top level planning goal rules for a family taking children to school scenario as indicated in the introduction is: 
\begin{verbatim}
Goal: in-school(children)
GP/PR rules:
in-school(children) <- frost | start(car,electric);
                               in-car(children,parent); 
                               drive(school)
in-school(children) <-       | in-car(children,parent); 
                               drive(school)
drive(school) <- late | drive(quick);park(close-school)
drive(school) <-      | drive(careful);park(safe)
\end{verbatim}
The first rule states that whenever the precondition ``frost" holds, the goal\\ ``in-school(children)" can be dealt with by the ``plan":\\ ``start(car,electric);in-car(children,parent); drive(school)''.\\ The second rule states that ``in-school(children)'' can alternatively be handled by the plan\\ ``in-car(children,parent); drive(school)''.\\
This specification makes use of the fact that in 2APL, rules are tried in order, and the first one that is applicable is executed. Thus the agent first checks whether there was frost that night and in case there is, it will start the car using an external electric connection and get the kids in the car and drive to school afterwards. If there was no frost, the condition of the first rule does not apply and the agent follows the second rule: ''in-car(children,parent); drive(school)''. A similar process appears for the ``drive(school)'' rules. It first checks the time. If they are late he will drive quick and park close to school. If they are not late, it will try the second rule and drive carefully and park safely.

A problem with this approach is that if the first rule fails during the execution of the plan, e.g. because the cable cannot be attached, the agent will try the next rule and start putting the kids in the car, even though the car did no start yet (and will not start on the battery). This can be avoided by explicitly indicating a precondition to be true or false in order to distinguish the different cases, as in the following two rules:
\begin{verbatim}
in-school(children) <- frost | start(car,electric);
                               in-car(children,parent); 
                               drive(school)
in-school(children) <- not(frost)|in-car(children,parent); 
                                  drive(school)
\end{verbatim}
However, if in this situation the plan associated with the first rule fails, the second rule is not applicable because there is frost. In this situation, the agent would just stop, without getting the cildren to school, as it has no applicable plan to follow. Moreover, in the case of conditions involving several criteria, the number of rules would quickly increase such that the different combinations of conditions could be represented.

We are not claiming that the above cases could not in some way be represented in 2APL. However, the example highlights two aspects that are interrelated and mingled in the 2APL representation. The conditions of the rules function as a precondition of the plan in the rule. However, the same conditions are also used for rule selection. The latter necessitates the constructions shown above but can also lead to (unexpected) difficulties as indicated.

By using the idea of social practice we distinguish conditions that are needed for rule selection and preconditions of plans. Thus we do not incorporate the conditions in all the rules, but are checking the context conditions of the social practice first and given those conditions select a subset of the rules that are relevant for the situation. Thus the deliberation is no longer purely goal driven, but is goal plus context driven. This will lead to a more natural specification and (through the modularization of rules based on context) to a more efficient deliberation.

\subsubsection{Case-Based Reasoning}
Whereas agent programming follows a goal-based approach for selecting actions, Case-Based Reasoning (CBR) is an example of reactive deliberation. CBR uses previous cases (or situations) as the basis for the selection of the next action~\cite{richter2013case}. The general cycle of CBR follows the following steps:
\begin{enumerate}
\item problem formulation
\item retrieve
\item reuse
\item revise
\item retain
\end{enumerate}
The first step is to formulate the problem. This is important because the way a problem is formulated will determine the query on the case-base, through which the most relevant case is eventually selected. In our scenario this might lead to the following (simplified) problem formulation:
\begin{verbatim}
frost = no
time = 8:30
car(fuel) = 10L
traffic = high
temperature = 6C
...
\end{verbatim}
For simplicity, we use here a very simple attribute-value structure, but more complex structures can also be used. With this formulation the case-base is searched for a similar situation. The likelihood of finding a case exactly like the current situation is minimal. Thus one needs some metrics in order to find the most `similar' case. Without getting into details, we just point to some difficult aspects here. Suppose there is an almost identical case except that the time is 8:50 (you are late). Could we use that case as a basis for the current course of action? So, even if the case in the case-base differs in only one parameter it might lead to a quite different course of action.

Given a case from the case-base that is close to the present situation, it has to be checked whether the plan for that situation can be used as it is or should be adjusted. E.g. if in the case from the case-base the first step would be to start driving very quick we can now relax and drive more carefull. Although for humans it is reasonably obvious how to make such a revision of the plan it is more difficult to find an algorithm that could calculate the necessary adjustments automatically.
Finally, the system should decide whether the present case and its course of action are sufficiently different from the cases in the case base to warrant adding it (in the right place).

Intuitively the example makes clear that even in simple scenarios like taking the children to school there are many parameters that potentially influence the course of action and even small differences can have big consequences for the course of action to be followed. Thus one needs to have a very large case base to cover all relevant cases such that an appropriate course of action is followed in each situation. In many domains (like crisis management) such a large case base cannot easily be assembled nor is it possible to construct one on the fly, because the consequences of errors are too big to allow for a gradually improving system.

\subsubsection{Work Practice Simulation}
Another approach relevant to social deliberation is that advocated by the Brahms platform~\cite{sierhuis09}. Brahms is a multi-agent, rule-based, activity programming language. The Brahms language allows for the representation of situated activities of agents in a geographical model of the world. Situated activities are actions that happen in the context of a specific situation, thus their execution is constrained not only by the reasoning capabilities of an agent, but also by the agent's beliefs of the external world, such as where the agent is located, the state of the world at that location and elsewhere, located artefacts, activities of other agents, or communication with other agents and artefacts.

The philosophy of Brahms comes from the realization that work practices in organizations differ from the work flows as described and prescribed by the organization. If it is recognized that ultimately employee behaviours, rather than management practices, are the key to success in organizations~\cite{colvin07}, then these practices should be described as agent behaviours rather than the official (goal directed) plans. Within Brahms a work practice is defined as the (collaborative) performance of collective situated activities of a group of people who collaborate and communicate, while performing these activities synchronously or asynchronously, by making use of knowledge previously gained through experience in performing similar activities. Differences between formal plans and the work practice can lead to unforeseen results and render organizational plans useless.

The Brahms modelling language is geared towards modelling people's activity behaviour \cite{sierhuis09}.
The Brahms framework consists of several interrelated models. Of particular relevance for this paper is the Activity Model that defines the behaviour of agents and objects by means of activities and workframes. Brahms has an activity-based subsumption architecture by which an agent's activities can be decomposed into sub-activities. Activities can be interrupted and resumed, just as humans can multi-task by switching between different activities. Workframes control when activities are executed based on the beliefs of the agent, and on facts in the world. However, as in CBR, workframes require the full instantiation of all its preconditions in order to be applied.

Our scenario could be modelled through the use of several workframes. One for starting the car, one for getting the children in the car, one for driving to school and one for parking the car. By giving the starting the car framework a high priority it will try to execute first. Thus if the preconditions of the workframe are fulfilled (which will include the presence or absence of frost) it will start the car using the start motor and normal battery. If the workframe cannot be executed the one with the next highest priority will be executed, etc. If during the starting of the car it is discovered that there was frost and the car does not start the framework for starting the car on an external electric supply is automatically fired and the start car workframe is interrupted. This makes the Brahms framework quite flexible. However, it has the same problem as the 2APL framework in that the context and preconditions of the workframes are mixed. Moreover, the agents also do not have a learning capability that might lead to a priority adjustment of workframes after the (failed) execution of activities.
\subsubsection{Agent organizations}
A last approach very relevant for social deliberation is the use of explicit agent organizations. See \cite{OperA} for a good description. Agent organizations also decsribe a context in which the agents are operating and which guarantees certain behavior. However, agent oragnizations do not specify the ``materials" used in their operation as do social practices. I.e. organizations do not specify specific times, places and objects available in the context of an interaction. The agent organization can be seen as a top-down abstract (social) structure within which agents operate, while a social practice is an emerging abstraction from the actual repeated interactions that agents perform. Having said this, there are many similarities between elements of social practices and agent organizations as well. Both have a kind of expected workflow, roles and meanings of actions in that context. We will later on use these similarities in the more formal definitions of social practices and also point out where the exact diferences lay.

\subsubsection{Conclusions}
It has become clear that, although there are several techniques from AI and computer science that seem appropriate to model social practices, they all fall short. Thus in the rest of this paper we investigate what are the main characteristics of social practices that need to be modeled, how they can be made more precise and create a formalization that can be used for implementing social practices.

\section{Characteristics of Social Practices}\label{characteristics}
As we have stated in the previous section, in Social Sciences, \emph{social practices} are defined on the basis of materials, meanings and competences~\cite{holtz2014}. 
Although social practices provide a handle for modelling the deliberation of socially intelligent agents because they seem to combine the elements that we require for socially intelligent behaviour, they are a relatively novel and not precisely defined concept from sociology that cannot be easily formalized in one step without making many implicit assumptions. Therefore we first make a step to get the notions involved in social practices more concrete. We are aware that such a step can already exclude some features that people might consider crucial for social practice theory. However, by making this step explicit, we can more easily adapt the concrete structure and add other elements or change them based on the experiences. The following is an adaptation of an initial conceptualization of social practices, as introduced in  \cite{Dignum2015}:\\
\textit{\textbf{Context}}
    \begin{itemize}
    \item \textit{Roles} describe the competencies and expectations about certain types of actors~\cite{sunstein96}. Thus a parent is expected to be able to drive the car to school and/or work.
    \item \textit{Actors} are all people and autonomous systems involved, that have capability to reason and (inter)act. This indicates the agents that are expected to fulfill a part in the practice.
    \item \textit{Resources} are objects that are used by the actions in the practice such as car, coats, school bags, etc. So, they are assumed to be available both for standard actions and for the planning within the practice.
    \item \textit{Affordances} are the properties of the context that permit social actions and depend on the match between context conditions and actor characteristics~\cite{gaver1996situating}.E.g. a sweater might be used (sometimes) as a coat because it also affords warmth.
    \item \textit{Places} indicates where all objects and actors are usually located relatively to each other, in space or time: The car is in the garage, school bags in the rooms of the children or in the corridor, etc.
\end{itemize}
\textit{\textbf{Meaning}}
    \begin{itemize}
    \item \textit{Purpose} determines the social interpretation of actions and of certain physical situations. E.g. the purpose of taking the children to school is that they can learn (or just to avoid a possible penalty for keeping children out of school)
    \item \textit{Promotes} indicates the values that are promoted (or demoted, by promoting the opposite) by the social practice. Driving carefully promote the value of universalism, because you take care of all passenger's safety and the safety of other road users. 
    \item \textit{Counts-as} are rules of the type "X counts as Y in C" linking brute facts (X) and institutional facts (Y) in the context (C) \cite{searle1995construction}. E.g., in a voting place, filling out a ballot counts as a vote.
    \end{itemize}
\textit{\textbf{Expectations}}
    \begin{itemize}
    \item \textit{Plan patterns} describe usual patterns of actions~\cite{bresciani2004tropos} defined by the landmarks that are expected to occur.
    \item \textit{Norms} describe the rules of (expected) behavior within the practice. They are decribed as deontic rules using obligations, prohibitions and permissions.
    \item \textit{Strategies}  indicate possible trigger-action combinations that can be performed at any time within the practice. Not all strategies need to be performed! They are meant as potential courses of action, whenever the trigger is satisfied. Strategies are specified as AIC statements. E.g. once all children are in the car the parent starts off for school. 
    \item \textit{Start condition}, or trigger, indicating how the social practice starts. At 8:30 the social practice of taking children to school starts.
    \item \textit{Duration}, or \textit{End condition}, indicating how the social practice ends. E.g. taking the children to school normally takes around 20 minutes and ends when the children are at school.
    \end{itemize}
\textbf{\textit{Activities}}
    \begin{itemize}
    \item \textit{Possible actions} describes the actions that usually occur in the social practice. This gives a kind of repository of actions with which activities in the social practice can be planned.
    \item \textit{Requirements} indicate the type of capabilities or competences that the agent is expected to have in order to perform the activities within this practice.
    \end{itemize}

\begin{figure*}[ht]
\centering
\includegraphics[width=0.9\textwidth]{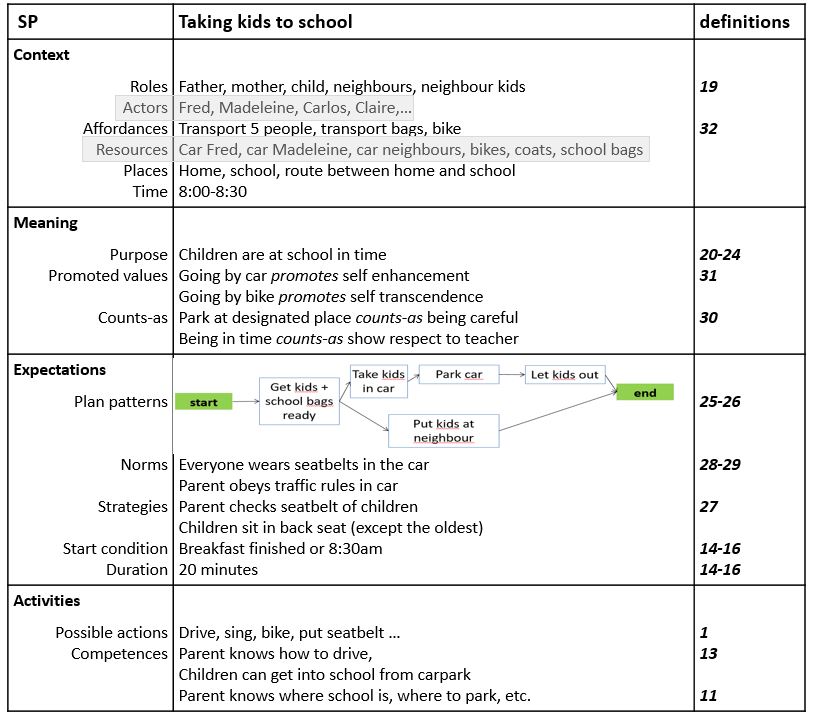}
\caption{social practice}
\label{fig:socialpractice}
\end{figure*}

An example of the social practice of taking the children to school is given in figure \ref{fig:socialpractice}. The last column of this table points to the formal definitions that are given in section \ref{sem} and that will be used to give each of these elements a precise, formal semantics in figure \ref{fig:SPform}.

The first block of a social practice consist of the physical and social context and is related to the definition of context given in ~\cite{Zimmermann:2007}. Two of the lines in the table are grey: actors and resources. In principle these lines pertain to what is available in a particular instance of the social practice rather than the general definition of the social practice.\\
The meaning of a social practice links concrete actions to social actions, values and purposes. We are aware that this part of the social practice might be extended with more elements like emotions, etc. However, not all social practices have these elements, while we claim that the elements we include are always part of any social practice.\\
The expectations are the most central part of the social practice as they pertain to the expected sequence of events that each actor expects to happen and on the basis of which she will plan and adapt behaviour.\\
The last block contains the prerequisites that actors are expected to have and the actions that one can expect to observe (plus their associated expected effects).\\
As said before Social Practices show some resemblance to agent organizations (see e.g. \cite{OperA}), in the sense that both provide structure to the interactions between agents. However, organizations provide an imposed (top-down) structure, while the social practices form a structure that arises from the bottom up. Thus instead of forcing compliance, interaction patterns in a social practice indicate expectations on the behavior of its actors. Therefore no guarantee is given that behavior will occur (exactly) as expected. This has a large influence on the way the social practices are formally specified, focusing on possibilities and priorities rather than prescribing behavior. 

Expectations are tied to the roles that determine possible and expected behavior and also indicate who is supposed to take initiative at certain points. E.g. the parents are supposed to take the initiative to start taking the children to school. Norms determine the normal patterns of behavior and thus also determine a certain type of expectation, namely the patterns of actions that are allowed or prohibited. The plan patterns also partly determine the sequences of actions that are expected. The different phases are temporally ordered and have a specific starting and end situation. Thus the whole practice should be fulfilled by following a trace that fits through that plan pattern.

Having this more concrete description of a social practice, we are now ready to develop a formal representation for social practices in the next section.

\section{Formal logic}\label{formal}
Taking the social science pespective on social practices seriously means that social practices have an existence independent of the individuals that execute them. In that respect they are like organizations, which also recruit persons to execute the functions belonging to specific roles and that provide workflows that partially (but not completely!) indicate which actions are expected from each individual. Given these type of similarities the formal model proposed in this paper is based on modal logics, extending work on agent organizations, in particular the Logic of Agent Organizations (LAO) \cite{Dignumigpl} that provides a formal, provable representation of organizations based on the notions of capability, `bring it about' -- or stit -- \cite{Porn1974} and attempt, indicating that an agent tries but might not always succeed. However, where LAO was based on a temporal logic and the stit operator, we will use dynamic logic with explicit actions for social practices. The difference is caused exactly by the fact that organizations are a top down structure where the results are what counts and the exact way they are achieved are left to the individuals. Social practices are emerging bottom up from repeated interactions between individuals using actions. These interactions then get abstracted into abstract social practices. This is reflected in the formalism that we use for scoial practices as well.\\
The semantics and model we give for the dynamic logic is quite complex and might seem at first sight overly complex. However, the complexity is needed to model two important features of actions in the context of social practices. First, we want to be able to state that a particular agent performs an action with a particular result, leaving open whether other agents performed any action at the same time. So, actions (state changes) are not exclusive. E.g. Fred takes the children to school does not preclude Madeleine to go to work at the same time. Secondly, we want to be able to express that a set of agents perform an action together, which entails that they all do some part of that action at the same time. E.g. Fred and the children go to school in the car together. We need to provide for these joint actions in the semantics as well.

Because the description of actions and action sequences is important as well as epistemic/doxastic states of the agents we will use a combination of dynamic logic with epistemic logic as the basis for our formalization.\footnote{Given that we only need to represent epistemic and dynamic operators, but not the epistemic changes, Dynamic epistemic logic \cite{Ditmarsch}, which is designed to capture the change of epistemic models by communication is not used. We merely need a logic in which both modalities are available.}

\subsection{Language}

The alphabet consists of a set $P$ of propositional symbols (p),
the set of violation symbols $V_i$ (indicating the violation of a norm), the operators $DONE$ and $DO$, a set
of agent identifiers $Ag$ (groups of agents identifiers are denoted
by $A, B, \ldots$), a set $Act$ of atomic action symbols typically
denoted by $\underline{a}$ 
%(this set at least includes the coordination actions $coordinate(a_i,\alpha_i,Y)$ with $a_{i} \in Ag$)
, the doxastic operators $B$ and $CB$, and the dynamic operators $[ \ ]$ and $\left\langle
\ \right\rangle$. The language $\mathcal{L}$ is based on three
types of syntactic constructs: \\
The set $ActExp$ of action expressions ($\alpha$) is defined through the following BNF:
\[
\alpha ::= \underline{a} \ | \ skip \ | \ \overline{\alpha} \ | \ \alpha_{1} +
\alpha_{2} \ | \ \alpha_{1} \& \alpha_{2} \ | \ \alpha_{1} ; \alpha_{2}.
\]
where $skip$ represents a ''doing nothing" action, $\overline{.}$
stands for the negation operator, $+$ stands for the
indeterministic choice operator, $\&$ for the parallel performance
operator and $;$ for the sequencing operator. We sometimes refer to the set of action expressions of the form $\underline{a}$, $skip$ or $\overline{\underline{a}}$ as the set of \emph{basic} actions. We use $act(\alpha)$ to denote the set of basic actions that are part of the complex action $\alpha$.\\
In order to express the fact that an action is performed by an actor or set of actors we define a so-called set ($Evt$) of event expressions ($\xi$) is defined through the following BNF:
\[
\xi ::= A:\alpha \ | \ \overline{A: \alpha} \ | \xi_{1} + \xi_{2} \ | \  \xi_{1} \&
\xi_{2} \ | \ \xi_{1} ; \xi_{2}.
\]
So, $A:\alpha$ denotes that the set $A$ of actors performs the action $\alpha$ together. In the same way $\overline{A: \alpha}$ denotes the negation of $A$ performing $\alpha$ together. (The exact semantics of this negation will be described in section \ref{sem}). Notice that the same notation for actions and event operators
(negation, $+, \& , ;$) is used. Nevertheless they belong to different categories of operators! We chose,
however, to keep notation as simple as possible as we have to introduce many different layers of operators later on already. 
The set $Ass$ of assertions ($\phi$) of our language $\mathcal{L}$ is defined through the following BNF:
\begin{definition}{{\bf (Language)}}\mbox{}\\
\begin{center}
\begin{tabular}{r|l}
 $\phi ::=$&  $p \ | \ V_i \  | \ DONE(\xi) \ | \ DO((\xi) \ | \ \neg \phi \ | \ \phi_{1} \vee \phi_{2} \ | \ \phi_{1} \wedge \phi_{2} \ | \ \phi_{1} \rightarrow \phi_{2} \ |$ \\
&  $Cap(a,\alpha) \ | \ [\xi] \phi \ | \ \langle\xi\rangle\phi \ | \ B_{i}(\phi) \ | \ EB_{A}(\phi) \ | \ CB_{A}(\phi) \ | \ Goal_i(\phi)$\\
\end{tabular}
\end{center}
\end{definition}
Thus we have all the usual propositional logic formulas and the violation symbols. We use $DO(a:\alpha)$ (resp. $DO(A:\alpha)$, $DONE(a:\alpha)$, $DONE(A:\alpha)$) to denote that actor $a$ performs action $\alpha$ next (resp. the set of actors $A$ performs action $\alpha$ (jointly), the actor $a$ performed action $\alpha$ as the last action or the set of actors $A$ jointly performed the action $\alpha$ as the last action).
We can refer to the capabilities of an actor $a$ through $Cap(a,\alpha)$ that is true if actor $a$ has the capability to perform $\alpha$. $[\xi]\phi$ indicates that $\phi$ will always be true after $\xi$ has happened. $\langle\xi\rangle\phi$ indicates that at least in one state after $\xi$ happened $\phi$ will be true. We can express doxastic statements as well, $B_i(\phi)$ states that $i$ believes $\phi$. $EB_A(\phi)$ states that everyone in the set of actors $A$ believes $\phi$ and $CB_A(\phi)$ states that it is common belief among the set of actors $A$ that $\phi$ is true. We also have the goals of an agent $i$ defined by $Goal_i(\phi)$\\
It will be handy to refer to abstract actions such as "an action that achieves $\phi$" or "an action performed by $A$ that achieves $\phi$". For this we introduce the following definitions:\\
\begin{definition}{{\bf (Achieving)}}\mbox{}\\
Let $\alpha_i \in ActExpr$\\
Let $X=\{\alpha_i : [\alpha_i]\phi\}$ then $\alpha\phi=+_{\alpha_i\in X}\alpha_i$\\
Let $Y=\{A:\alpha_i : [A:\alpha_i]\phi\}$ then $\alpha(A)\phi=+_{A:\alpha_i\in Y}A:\alpha_i$
\end{definition}
Note that if $\phi=T$ (Where $T$ stands for the logical constant $true$) the abstract action refers to any action that can be performed in a particular state. Also note that the sets $X$ and $Y$ are dependent on the state in which they are evaluated. I.e. the set of actions that achieve a certain formula $\phi$ depends on the state where one starts performing those actions. Also, these sets $X$ and $Y$ can in principle be infinite large sets. However, if we keep the amount of operators in the expression finite (which in practice they are) these sets are also finite.
Given the above language, we now proceed to define the formal semantics for the constructs of that language.

\subsection{Models}

In order to give a semantics to the language introduced above we
start defining the notion of model for $\mathcal{L}$.
\begin{definition}{{\bf (Kripke Model)}}\mbox{}\\
{\rm A model $M$ is defined as follows:
\[
M = \left\langle \mathcal{P}^{+}(\mathrm{Agt}), \mathrm{Act} \cup \mathrm{skip}, \mathrm{Capability},
\mathbb{W}, \llbracket \ \rrbracket_{R}, \{ \mathbb{R}_{a} \}_{a \in \mathrm{Agt}}, \{ \mathbb{G}_{a} \}_{a \in \mathrm{Agt}},
\prec , \pi \right\rangle
\]
where:
\begin{itemize}
        \item $\mathcal{P}^{+}(\mathrm{Agt})$ is the non-empty powerset of the finite set of actors
$\mathrm{Agt}$, that means the possible groups of actors. We assume $\mathrm{Agt} = Agt$.
        \item $\mathrm{Act}\cup \mathrm{skip}$ is the set of actions. (Note these are semantic entities and different from the syntactic action expressions ($ActExpr$)).
        \item $\mathrm{Capability}$ is a function in $Agtx2^{Act}$ that indicates for each actor the set of (atomic) actions it is capable to perform.
        \item $\mathbb{W}$ is the set of possible states.
        \item $\llbracket \ \rrbracket_{R}$ is a function $f$ s.t. $f : Evt \times \mathbb{W} \longrightarrow \mathcal{P}(\mathbb{W})$, to each event expression-world couple it associates the set of states to which the performance of that event in that world leads. It consists of a composition of the two functions $\llbracket \ \rrbracket$ and $R$ which will be introduced in Section \ref{sem}.
        \item $\{ \mathbb{R}_{a} \}_{a \in \mathrm{Agt}}$ is a family of serial, symmetric and transitive accessibility relations which are indexed by actors indicating the believable worlds of agent $a_i$.
        \item $\{ \mathbb{G}_{a} \}_{a \in \mathrm{Agt}}$ is a family of serial and transitive accessibility relations which are indexed by actors indicating the worlds agent $a_i$ wants to get to.
        \item $\prec$ is a serial, partial ordering on $\mathbb{W}$ denoting the order in which worlds are reached through actual performances of events. This ordering is constrained as follows: if $w_{1} \prec w_{2}$ and $\exists w_{3}$ s.t. $w_{3} \mathbb{R}_{a} w_{1}$ or $w_{3} \mathbb{R}_{a} w_{2}$ then $w_{3} \mathbb{R}_{a} w_{2}$ and $w_{3} \mathbb{R}_{a} w_{1}$
%\footnote{This condition is important for proving validity (10) in Proposition \ref{validities}.}. 
From an intuitive point of view, this condition guarantees that if $w_1$ or $w_2$ is a believable world from $w_3$ than the other one should also be believable and thus the whole path of actual performances through $\mathbb{W}$ should be doxastically accessible.
        \item $\pi$ is a usual truth function $f$ s.t. $f: Ass \times W \longrightarrow \{ 1, 0 \}$.
\end{itemize}}
\end{definition}
Like in \cite{meyer88different,dignum90negations,dignum96modal}
our semantics consists of two parts: first event expressions are
interpreted as set theoretic constructs on $\mathrm{Act}$ where
events get a so-called {\em open} interpretation; successive
event expressions are interpreted as state-transition functions
determining the accessibility relation $\llbracket \
\rrbracket_{R}$ on $\mathbb{W}$.

\subsection{Synchronicity sets, steps, synchronicity traces, and worlds} \label{sem}

The interpretation of events is based on the basic notion of {\em
synchronicity set} (s-set). A synchronicity set indicates all basic events that are performed in parallel by a set of agents in a certain state. 
\begin{definition}{{\bf (s-set)}} \\
{\rm The set $\mathcal{S}$ of s-sets is defined as follows: $\mathcal{S} =
\mathcal{P}^{+}(\mathrm{Agt}) \times \{ \mathrm{skip} \} \cup
\mathcal{P}^{+}(\mathrm{Agt}) \times \mathcal{P}^{+}(\mathrm{Act})$.}
\end{definition}
Synchronicity sets, that is elements of $\mathcal{S}$, are denoted
by $S_{1}, S_{2}, \ldots$. The s-sets formalise the aforementioned open interpretation view on events.
Based on the notion of s-set we define the notion of {\em step}
\footnote{Notice that in \cite{dignum96modal} s-sets are called
steps, and no notion of step as it will be defined in this work
occurs there.}.
\begin{definition}{{\bf (Step)}} \label{step} \\
{\rm The set $Step$ of steps is defined as follows:
\begin{eqnarray*}
Step & = & \{ \times_{A \in \mathcal{P}^{+}(\mathrm{Agt})} \mathcal{S}_{A} \ | \
\forall A, B \in \mathcal{P}^{+}(\mathrm{Agt}) : B \subseteq A \Rightarrow act(S_{B})
\subseteq act(S_{A}) \ \&  \\
& & \forall A, B \in \mathcal{P}^{+}(\mathrm{Agt}) :  action(S_{B}) = \mathrm{skip}
\Rightarrow action(S_{A \cup B}) = action(S_{A}) \}
\end{eqnarray*}
where $action$ is a function that extracts the action component from a given s-set
($action(A: \{ a_{1}, a_{2} \})= \{ a_{1}, a_{2} \}$).}
\end{definition}
Steps represent a sort of snapshot of the activity of each
subgroup of $\mathrm{Agt}$ at a certain moment, depicting how all
agents move one ''step" ahead. Steps are therefore sets of s-sets
of cardinality $2^{n}-1$ where $n$ is the number of agents in
$\mathrm{Agt}$. They are constrained in such a way that whatever
action is performed by a subgroup is also performed by a
supergroup, and subgroups remaining inactive are treated as
performing a $\mathrm{skip}$ action. Steps, that is elements of
$Step$, are denoted by $s_{1}, s_{2}, \ldots$.

In order to provide a semantics for sequential expressions the
concept of {\em synchronicity trace} (s-trace) is needed. Notice
that this concept uses steps instead of s-sets like in
\cite{meyer88different,dignum90negations}.
\begin{definition}{{\bf (s-trace)}}
{\rm The set $\mathcal{T}$ of s-traces is defined as follows:
\[
\mathcal{T} = \{ \left\langle s_{1}, ..., s_{n}, ... \right\rangle | s_{1}, ...,
s_{n}, ... \in \mathcal{S} \}.
\]
The length of an s-trace $t$ is denoted by $dur(t)$. We assume
$dur(t)$ to be finite.}
\end{definition}

An event will be interpreted as a set of s-traces. The range for
our interpretation of events is a set $\mathcal{E}$ such that
$\mathcal{E} = \mathcal{P}(\mathcal{T})$. Elements of
$\mathcal{E}$ (sets of s-traces) are denoted as $T_{1}, T_{2},
\ldots$ The length $dur(T)$ of a set $T$ is defined as $max \{
dur(t) | t \in T \}$.
We stipulate the length of the empty set to be $dur(\emptyset) = 1$.

We can now introduce the operations that constitute the semantic counterpart of our
syntactic operators.
\begin{definition}{\bf (Operations on events)}\\ \label{operations}
{\rm Let $T_{1}, T_{2} \in \mathcal{T}$:
\begin{eqnarray*}
T_{1} \circ T_{2} & = & \{ t_{1} \circ t_{2} \ | \ t_{1} \in  T_{1}, t_{2} \in T_{2}
\} \\
T_{1} \doublecap T_{2} & = & \bigcup \{ t_{1} \doublecap t_{2} \ | \ t_{1} \in
T_{1}, t_{2} \in T_{2} \} \\
T_{1} \doublecup T_{2} & = &  T_{1} \cup T_{2} - (\bigcup \{ t_{1} \doublecap t_{2}
\ | \ t_{1} \in  T_{1}, t_{2} \in T_{2} \ \ \mbox{and} \ \ t_{1} \neq t_{2} \})\\
\tilde{T} & = & \left\{ \begin{array}{ll}
\mbox{if} \ \ T \neq \emptyset, & \tilde{T} =\doublecap \{ \tilde{s} \ | \ s \in T \} \\
\mbox{if} \ \  T = \emptyset, & \tilde{T} = Step \end{array}
\right.
\end{eqnarray*}
Where:
\begin{itemize}
        \item $t_{1} \circ t_{2}$ is defined as follows: if $t_{1} = \left\langle  s_{1},
..., s_{n} \right\rangle$ and $t_{2} = \left\langle s_{1}', ...,
s_{m}' \right\rangle$ then, $t_{1} \circ t_{2} = \left\langle
s_{1}, ..., s_{n}, s_{1}', ..., s_{m}' \right\rangle$.
        \item $t_{1} \doublecap t_{2}$ is defined as follows:  
$t_{1} \doublecap t_{2} = \left\{ \begin{array}{ll}
t_{1} \ \ \mbox{if} \ \ t_{2} \in start(t_{1}) \\
t_{2} \ \ \mbox{if} \ \ t_{1} \in start(t_{2}) \\
\emptyset \ \ \mbox{otherwise}
\end{array} \right.$

The function $start$ associates to a given s-trace all its starting possible s-traces.
   \item $\tilde{t}$ is defined as follows: $\tilde{t} = \bigcup_{1 \leq n \leq
dur(t)} \left\langle s_{1}, ... , \tilde{s_{n}} \right\rangle$, where $\tilde{s}
= Step - \{ s \}$\footnote{Negation of sequences constitutes a delicate matter. For a deeper discussion of this issue we refer to \cite{dignum90negations}.}.
\end{itemize}}
\end{definition}
Intuitively, we want $\doublecup$ to yield the property: $a \equiv
a+a;b$ for event expressions, because the choice states that we know at least $a$ will be performed and after that it is can be anything (because the first choice does not specify anything after $a$). In order to establish this property
we cannot just use a union of the sets of s-traces representing $a$
and $a;b$ but have to do some "cleaning up" by subtracting
superfluous parts.

The semantics of events are obtained by means of a function
$\llbracket \ \rrbracket : Evt \longrightarrow \mathcal{E}$ such
that:
\begin{definition}{{\bf (Semantics of events)}}
\begin{eqnarray*}
\llbracket A:\underline{a} \rrbracket & = & \{s \ | \ S_{A} \in s, a \in act(S_{A}) \} \\
\llbracket \xi_{1} ; \xi_{2} \rrbracket & = & \llbracket \xi_{1} \rrbracket \circ
\llbracket \xi_{2} \rrbracket \\
\llbracket \xi_{1} + \xi_{2} \rrbracket & = & \llbracket \xi_{1} \rrbracket
\doublecup \llbracket \xi_{2} \rrbracket \\
\llbracket \xi_{1} \& \xi_{2} \rrbracket & = & \llbracket \xi_{1} \rrbracket
\doublecap \llbracket \xi_{2} \rrbracket \\
\llbracket \overline{\xi} \rrbracket & = & \tilde{\llbracket \xi \rrbracket} \\
\llbracket skip \rrbracket & = & \{ \mathrm{skip} \}.
\end{eqnarray*}
\end{definition}
The basic clause stipulates that the meaning of an atomic event consists of the set
of steps where that action at least is performed by that specific group of agents.

On the basis of this evaluation for events, an evaluation of
groups performing complex actions is obtained:
\begin{definition}{{\bf (Semantics of collective actions)}}\\
{\rm Let $A=A'\cup A''$ then}
\begin{eqnarray*}
\llbracket A: \alpha_{1}; \alpha_{2} \rrbracket & = & \llbracket
A': \alpha_{1} \rrbracket \circ \llbracket A'': \alpha_{2} \rrbracket \\
\llbracket A: \alpha_{1} + \alpha_{2} \rrbracket & = & \llbracket A': \alpha_{1} \rrbracket \doublecup \llbracket A'': \alpha_{2} \rrbracket \\
\llbracket A: \alpha_{1} \& \alpha_{2} \rrbracket & = & \llbracket A': \alpha_{1} \rrbracket \doublecap \llbracket A'': \alpha_{2} \rrbracket \\
\llbracket A: \overline{\alpha'} \rrbracket & = & \llbracket \overline{A: \alpha'}
\rrbracket.
\end{eqnarray*}
\end{definition}
Thus if we indicate that a group $A$ performs an action $\alpha$ we explicitly do not differentiate who performs what part of that action. If we want to be more specific we can indicate the specific group performing a sub-action with that sub-action directly.It should be noted that we do not have by definition that $A:\alpha_1 \circ A:\alpha_2 = A:\alpha_1\circ\alpha_2$ (with $\circ$ being any of the operators connecting actions). E.g. if Fred and the children drive to school they perform the whole action together, but Fred is doing the driving and the children doing the passive passenger actions:
\[
Fred:drive\& \{Marco,Claire\}:sit \not\equiv \{Fred,Marco,Claire\}:(drive\&sit)
\]

To connect this interpretation of events to a possible world
semantics a function $R: \mathcal{E} \times \mathbb{W}
\longrightarrow \mathcal{\mathbb{W}}$ is defined, which couples
events with state-transitions.
\begin{definition}{{\bf (Function $R$)}} \\
$R(T,w_{1}) = \{ w_{2} \ | \ \exists t \in T \ \ \mbox{s.t.} \ \ w_{2} = R(t, w_{1})\}$
{\rm where $R$ on transitions is inductively defined as follows:
\begin{eqnarray*}
R(s_{1}, w_{1}) & = & r(s_{1}, w_{1}) \\
R(t_{1} \circ t_{2}, w_{1}) & = & R(t_{2}, R(t_{1}, w_{1})).
\end{eqnarray*}
and $r: \mathcal{S} \times \mathbb{W} \longrightarrow \mathbb{W}$, that is a
function that, given a state, returns the following state reachable through a given
synchronicity set, and such that $r(\{ \mathrm{skip} \}, w) = w$.}
\end{definition}

This concludes the complex semantics of actions and sequences of actions performed by individuals and groups. Now all ingredients are in place to evaluate formulas.

\subsection{Evaluating formulas}

The meaning of formulas $\phi$ in a world $w$, given the structure
$M$, is defined as usual. For space reasons we report here only the dynamic clauses, the belief clauses and the $DONE$ and $DO$ unary
operators.
\begin{definition}{{\bf (Semantics of assertions)}}\\
{\rm In the following let} $dur(\llbracket\xi_1\rrbracket)=1$,
\begin{eqnarray*}
M, w_{1} & \models & [ \xi ] \phi  \ \ \mbox{iff} \ \  \forall
w_{2} \in \llbracket
\xi \rrbracket_{R}(w_{1}), M, w_{2} \models \phi \\
M, w_{1} & \models & \left\langle \xi \right\rangle \phi  \ \ \mbox{iff} \ \
\exists w_{2} \in \llbracket \xi \rrbracket_{R}(w_{1}), M, w_{2} \models \phi \\
M, w_1 & \models & Cap(i,\alpha) \ \ \mbox{iff} \ \forall \underline{a}\in\alpha: a \in Capability(i) \\
M, w_{1} & \models & DONE(\xi_1) \ \ \mbox{iff} \ \exists w_{2} \in \mathbb{W}, \ w_{2} \prec w_{1} \Rightarrow w_{1} \in \llbracket \xi_1 \rrbracket_{R} w_{2} \\
M, w_{1} & \models & DONE(\xi ; \xi_1) \ \ \mbox{iff}\ \ \exists w_{2} \in \mathbb{W}, \ w_{2} \prec w_{1} \Rightarrow M, w_{1} \models DONE(\xi_{1}) \ \ \mbox{and} \\
& & M, w_{2} \models DONE(\xi)\\
M, w_1 & \models & DO(\xi_1) \ \ \mbox{iff}\ \ \forall w_2 \in \mathbb{W}, w_1 \prec w_2 \Rightarrow w_2 \in \llbracket \xi \rrbracket_R w_1\\
M, w_1 & \models & DO(\xi_1;\xi) \ \ \mbox{iff}\ \ \forall w_2 \in \mathbb{W}, w_1 \prec w_2 \Rightarrow w_2 \in \llbracket \xi \rrbracket_R w_1 \ \ \mbox{and} \\
& & M, w_2 \models DO(\xi)\\
M, w_1 & \models & B_a(\phi) \ \ \mbox{iff}\ \ \forall w_2 \in \mathbb{W}, \mathbb{R}_a(w_1,w_2) \rightarrow M,w_2\models \phi\\
M, w_1 & \models & EB_A(\phi) \ \ \mbox{iff}\ \ \forall a \in A, M,w_1\models B_a(\phi)\\
M, w_1 & \models & CB_A(\phi) \ \ \mbox{iff}\ \ M,w_1\models EB_A(\phi) \wedge_{i=2}^{\infty} EB_A^i(\phi)\\
M, w_1 & \models & Goal_a(\phi) \ \ \mbox{iff}\ \ \forall w_2 \in \mathbb{W}, \mathbb{G}_a(w_1,w_2) \rightarrow M,w_2\models \phi
\end{eqnarray*}
\end{definition}
Informally, a sentence $[ \xi ] \phi$ ($\left\langle \xi
\right\rangle \phi$) is true in $w$ iff $\phi$ is true in every
world (respectively in at least one world) accessible through a
performance of $\xi$.  An agent $i$ is capable of performing an action $\alpha$ if all the atomic actions that $\alpha$ is composed of are part of the capabilities of $i$. Sentences $DONE(\xi)$ are evaluated as true
in a world $w$ iff that world may be reached from $w_{n}$
performing $\xi$, and $\xi$ is actually performed in $w_{n}$ and
all intermediate worlds $w_{i}$ are reachable from $w_{n}$
performing starting sequences of $\llbracket \xi \rrbracket$ of
length $n-i$ along the $\prec$ ordering. The semantics of $DO(\xi)$ is very similar, but then forward directed. Note that if $DO(\xi)$ is true it does not imply that this is the only thing that is happening. It just means that $\xi$ will be done for certain.\\
The last assertions are about beliefs, everyone believes and common beliefs. Everyone in a group $A$ believes $\phi$ if they all individually believe $\phi$. There is a common belief in $\phi$ if they all believe in $\phi$ and also believe that they all believe in $\phi$ etc. These are the common definitions of the belief modalities (see e.g. \cite{meyer95epistemic}).\\
\subsection{Some additional constructs}
In addition to the above definitions we use $DO(a,\underline{b},\alpha(A))$ and $DONE(a,\underline{b},\alpha(A))$ to denote that $a$ performs action $\underline{b}$ as part of the group $A$ in order to perform $\alpha$ together. The definition of these constructs is given as follows.
\begin{definition}{{\bf (Acting as part of a group)}}\\
Let $a \in A$ and $A'=A\\a$ and let $\underline{b} \in act(\alpha)$ then\\
\begin{eqnarray*}
M,w \models DO(a,\underline{b},\alpha(A)) \ \mbox{iff} \ M,w\models DO(a:\underline{b}) \wedge DO(A:\alpha)\\
M,w \models DONE(a,\underline{b},\alpha(A)) \ \mbox{iff} \ M,w \models DONE(a:\underline{b}) \wedge DONE(A:\alpha)
\end{eqnarray*}
\end{definition}
Based on the above semantics we can also define a number of operators that are sometimes easy to use as they give different ways to abstract from an action or indicate that an action is possible or attempted.
Sometimes we want to refer to the fact that an agent is not just capable, but also able in the present state to perform an action $\alpha$, to achieve a certain result $\phi$, to attempt to achieve $\phi$ ($H_a\phi$) or to see to it that (stit) $\phi$ ($E_a\phi$). This is defined as follows.
\begin{definition}{{\bf (Ability, attempt and stit)}}\\
\[M, w\models G_a\alpha \;\mbox{iff}\; M,w\models Cap(a,\alpha) \wedge <a:\alpha>T \]
\[M, w\models G_a\phi \;\mbox{iff}\; \exists\alpha \ M,w\models Cap(a,\alpha) \wedge \langle a:\alpha \rangle\phi \]
\[M, w\models H_a\phi\; \mbox{iff}\; \exists\alpha \;M, w\models \;DO(a:\alpha) \wedge \langle a:\alpha \rangle\phi \]
\[M, w\models E_a\phi \;\mbox{iff}\; \exists\alpha\;M, w\models DO(a:\alpha) \wedge [a:\alpha]\phi \]
\end{definition}
These operators can be used to indicate that an actor does not only have the capability to perform an action, but the situation is such that she actually could perform it. So, this is an easy way to indicate that all preconditions of an action are fulfilled. The attempt operator is used in cases where the outcome of an action is depending on other actors or the environment. In a social practice it can be important that even though a result was not achieved it is known that it was attempted. Because that means that the actors tried to keep to the social practice, but just were unable to do their part. This can have different consequences than when actors just don't do what they were expected or ought to have done.
%These definitions make it also possible to define the temporal operator $NEXT$ in terms of our dynamic logic:
%\[M, w\models NEXT(\phi)\; \mbox{iff}\; M, w\models\exists a\in Agt:\; E_a\phi\; or\; M,w\models \phi \wedge \neg \exists a\in Agt:\; H_a\neg\phi\]

\section{Formalizing Social Practices}
\label{socialpracticeformal}
Given the formal logic developed in the previous section we can now give a formal account of the elements of social practices as we have sketched in section \ref{characteristics}. We should point out that we actually formalize the \emph{instances} of social practices. For some elements there will be no difference between the social practice and its instances (e.g. the meanings are the same for all instances of a social practice), but e.g. the start and end conditions refer to the conditions of a particular instance. In the next section we will get back to this issue and indicate how we go from the description of instances of social practices to the social practices themselves. Although the basics of the logic are all given we will introduce some new constructs when needed, but always based on the same Kripke models as already introduced. There are a lot of definitions needed to describe all elements in the logical formalism. Unfortunately, the logical definitions have to be built up from basic building blocks towards the more complex elements in the social practice and thus do not completely follow the parts as used in the description of social practices used in figure \ref{fig:socialpractice}. So, for ease of reference we have linked the definitions from this section to figure \ref{fig:socialpractice} such that it becomes clear where the definitions are needed for.

The first thing that has to be established when performing a social practice is that the social practice is ``active". I.e. the start condition was true and the end condition is not yet true. These two elements are part of the expectations of a social practice rather than the context of it. In the context of the social practice we specify things that can be assumed to be available during the social practice, while the start and end conditions define the boundaries of the social practice.
This will be especially important when effects of an action depend on the context. E.g. the effect of raising your hand can be different in an auction and in a classroom. Intuitively a context stands for a temporal and spatially defined  interval within which an action takes place, such as "the lunch break in the dining room of the school". 

There is a vast amount of literature on reasoning in context. We will not tap on this literature, because we mainly use contexts (in this paper) as reference structures needed to specialize interaction expectations and beliefs. Every social practice can also function as a context by using only those parts that give temporal and spatial restrictions on the actions and agents.\\
We extend the language $\mathcal{L}$ with a set $C=\{c_1,...,c_n\}$ of special context constants (or names). We use these constants to identify social practices as well. Thus, let $SP$ be the set of social practice identifiers then: $SP \subset C$. (Thus every social practice is a context, but not every context is a social practice). 
We have the following definitions on contexts:
\begin{definition}{{\bf (Context)}}\\
\[ \llbracket c \rrbracket \equiv U \subset \mathbb{W} \]
{\rm such that $connected(U,\prec)$, i.e. $U$ is a connected graph using $\prec$ as links between the worlds.} 
\end{definition}
We explicitly do not have closedness of this set under combinations of contexts in any way! We also do not assume that any combination of specifications of temporal and spatial intervals defines a context.

Because $\prec$ defines a partial order on $U$ we can define: 
\begin{definition}{{\bf (Start and End of contexts)}}\\
{\rm Let $U$ be the set of worlds associated with a context $c$ then}
\[ start(U)=\{w | \neg \exists w'\in U: w'\prec w\}\]
\[ end(U) =\{w | \neg \exists w'\in U: w\prec w'\}\]
\end{definition}
With these definitions we can now define the start and end conditions of a social practice. 
\begin{definition}{{\bf (Start Condition (SC) and End Condition (EC))}}\\
{\rm If $sp_i$ is a social practice then}\\
\begin{center}
\begin{tabular}{ll}
 $M,w \models$ & $SC(sp_i,\phi) \ \mbox{iff}$\\
 &$\forall w'\in start(\llbracket sp_i \rrbracket) M,w'\models \phi \ {\rm and} \ \forall w''\not\in start(\llbracket sp_i \rrbracket) M,w'\models \neg\phi$\\
$M,w \models$ & $EC(sp_i,\phi) \ \mbox{iff}$\\
&$\forall w'\in end(\llbracket sp_i \rrbracket) M,w'\models \phi \ {\rm and} \ \forall w''\not\in end(\llbracket sp_i \rrbracket) M,w'\models \neg\phi$
\end{tabular}
\end{center}
\end{definition}
Intuitively this definition just states that the start condition $SC$ should hold at the start (and only at the start) of the social practice and the end condition $EC$ at the end (and only at the end) of the social practice. We also define that a social practice is active which is handy later on.
\begin{definition}{{\bf (Active Social Practices)}}\\
\[ M, w \models active(sp) \ \mbox{iff} \ w\in \llbracket sp \rrbracket\] 
{\rm Let $A_{sp}$ be the set of actors in $sp$ then we have the following axiom}
\[ M, w \models active(sp) \rightarrow CB_{A_{sp}}(active(sp))\]
\end{definition}
I.e. when a social practice is active the actors involved in it believe that it is active. The set $A_{sp}\subset Agt$ is defined for each context as the set of actors that is involved. Not necessarily every actor needs to perform an action. Some actors might just be observing or receiving messages.

In general it can be the case that more than one context is active when an actor performs an action. This is not always a problem, but especially for concepts like expectations we want to have a unique context to narrow down (disambiguate) which expectations are relevant. Therefore we also define a function $Salient(a:\gamma,c_i)$.
\begin{definition}{{\bf (Salient)}}\\
\[ M,w\models Salient(a:\gamma,c_i) \ \mbox{iff} \ w\in \llbracket c_i \rrbracket \ \mathrm{and} \ \neg\exists c_j: w\in \llbracket c_j \rrbracket \ \mathrm{and} \ \llbracket c_j \rrbracket \subset \llbracket c_i \rrbracket \]
\end{definition}
which states that the most salient context that is active for actor $a$ when performing $\gamma$ is the smallest (most specific) context in which it is performed.

Before defining the action related elements of the social practice we first will extend the language with a possibility to talk about roles and actors playing roles. We follow \cite{Dignumigpl} and extend the language $\mathcal{L}$ with $play(a,r,c)$ to indicate that actor $a$ enacts role $r$ in context $c$. E.g. Fred can play the role of driver when taking he children to school. Fred is still also playing the role of father in the larger context of the family life. So, an actor can play several roles at the same time and a role can be played by several actors. E.g. there are several children that play the role of pasenger in the car. We can use $play(a,r)$ when we refer to $a$ playing the role $r$ from the salient context $c$, which does not have to be mentioned.\\
The model $M$ is also extended with a relation $rea(a,r,w)$ that relates the actors with the roles they play in a certain state $w$. 
The semantics of the the construct are given as follows.
\begin{definition}{{\bf (Role playing actors)}}\\
{\rm We extend the model M with a set of role names for each context $c$: $RR_c$ and a relation $rea$ that connects actors to roles in a world, with the constraint that the relation can only exist if the world is part of the context for which the role is defined.}
\[M,w\models play(a,r,c) \ \mbox{iff} \ r\in RR_c \wedge w\in\llbracket c \rrbracket \wedge rea(a,r,w)\]
\end{definition}
Having roles defined makes it possible to talk about the set of agents playing a role. Therefore we can now use the following abreviation. Let $r\in RR_c$ then
\[ r:\gamma \ \equiv \ \exists A\subset \{a|play(a,r,c)\} \  {\rm and} \ A:\gamma \]
So, we can say that a role performs an action, meaning that some subset of the actors playing that role perform the action. Thus, in all places where we use abstract actions, we can now also have that a role is performing such an abstract action. It allows for a step between a very specific group of actors performing an action and any group of actors performing that action. This is something that appears very often in social practices, because many actions are supposed to be taken by some actor playing a particular role. E.g. the parent is supposed to drive the car with the children to school. It does not matter which parent it is, but it should be one of them and not a child.

Based on all the previous definitions we now start to define the purpose of a social practice and plan patterns of a social practice. We start with the purpose of an action (and a social practice). The intuitive meaning of the purpose of an action is the reason for which the action is performed in the social practice. It is part of the intended and actual result of the action and thus indicates that the action is performed with the \emph{goal} of achieving that part of the result. E.g. when a lecturer raises her hand the students stop talking, which is the purpose of raising her hand. However, after the action her hand is also raised, which was intended, but not the purpose of the action.
\begin{definition}{{\bf (Purpose of a basic action)}}\\
{\rm The purpose of a basic action $\alpha$ performed by $a \in A_c$ in a context $c$, where $A_c$ is the set of all actors present in $c$ is defined as}\\
\begin{center}
\begin{tabular}{l} 
$M,w\models purpose(a:\alpha,c,\phi)$ \mbox{iff} \\ 
$M,w\models CB_{A_c} ((Salient(a:\alpha,c) \wedge DO(a:\alpha))\rightarrow Goal_a(\phi) \wedge B_a([a:\alpha]\phi))$
\end{tabular}
\end{center}
\end{definition}
Note that $a$ needs to perform $\alpha$, have a belief that $\alpha$ leads to the achievement of $\phi$ and also has this $\phi$ as its goal. Thus we take the meaning of a purposeful action to be that the consequence of that action was intended. In top of all that, $a$ does this particular action in a context, in the presence of other actors. This means that the fact that $a$ does action $\alpha$ is not just an individual thing anymore, but is a common belief among all the actors in that context. This actually says that all actors in a social practice know what is going on and understand why actions are performed by the actors involved. E.g. when Fred drives the car (with the children) to school, he does it with the purpose of getting the children to the school. And both Fred and the children believe that this is his goal in this practice.

If we talk about the general purpose of an action $\alpha$ in context $c$ we mean that whenever any actor performs that action we infer that it was done to achieve that purpose. And the same if $\alpha$ is an abstract action $\beta\phi$ or $\alpha$=$+_{i=1...n}\alpha_i$
\begin{definition}{{\bf (Purpose of general actions)}}\\
\begin{center}
\begin{tabular}{l} 
$M,w\models purpose(\alpha,c,\phi)$ \mbox{iff} \\ 
$\forall a\in A_c: M,w\models CB_{A_c} ( (Salient(a:\alpha,c) \wedge DO(a:\alpha))\rightarrow Goal_a(\phi) \wedge B_a([a:\alpha]\phi))$
\end{tabular}\\
\end{center}
{\rm and if} $\alpha$=$\beta\phi$=$\cup_{i=1...n}\alpha_i$ {\rm then} $purpose(\alpha,c,\phi) \equiv \forall \alpha_i: purpose(\alpha_i,c,\phi)$
\end{definition}

We have similar definitions for the purpose of complex actions.
\begin{definition}{{\bf (Purpose of complex actions)}}\\
{\rm Let $\gamma \in ActExp$ then}\\
\begin{center}
\begin{tabular}{rl}
$M,w\models$&$purpose(a,\gamma,c,\phi)$ \mbox{iff} \\
$\forall \alpha_i\in action(\gamma): M,w\models$&$CB_{A_c} (Salient(a:\gamma,c) \wedge DO(a:\alpha_i)\rightarrow$\\ 
&$ Goal_a(\phi) \wedge B_a([a:\gamma]\phi))$
\end{tabular}
\end{center}
\end{definition}
Note that this definition only indicates that the whole sequence $\gamma$ leads to $\phi$ and does not restrict exactly which actions are part of that sequence. E.g. The children should be at school before 9am. It does not matter which parent takes them, which car they take, or if they go by bike. However, every action is believed to be done because $a$ has the goal to achieve $\phi$. In this way we represent the intuition that each action of the complex action $\gamma$ contributes to the goal $\phi$.\\
The idea of the above purposes is close to the notion of landmarks. The use of landmarks in the specification of agent plans and interactions has been proposed by \cite{kumar02,Dignum2007}. Formally, landmarks are conjunctions of logical expressions that are true in a state, representing families of protocols. The level of landmark detail determines the degree of actors freedom. In our context we use purposes instead of landmark, because the state is not just a predetermined state that has to be achieved, but it also defines the reason for the interaction and thus drives the commitments, planning, etc, of the participating actors.\\
We can further generalize the definition of purpose first by allowing the different actions in $\gamma$ to be done by different actors $a_j$ from a set of actors $A_c$. And we can also abstract altogether from the agency and give the purpose of the complex action $\gamma$. E.g. checking the minutes of the previous meeting at the start of a formal meeting is done for the purpose of creating common ground and agreement about what happened in last meeting.
\begin{definition}{{\bf (Purpose of group and role actions)}}\\
{\rm Let $act(\gamma)=\{\alpha_1;...;\alpha_n\}$ and $A_c=\{a_1,...,a_m\}$ and $r\in RR_c$ then}\\
\begin{center}
\begin{tabular}{l}
$M,w\models purpose(A_c,\gamma,c,\phi)$ \mbox{iff} \\ 
$\forall \alpha_i\exists a_j; M,w\models CB_{A_c} (Salient(A_c:\gamma,c) \wedge DO(a_j:\alpha_i)\rightarrow Goal_{a_j}(\phi) \wedge B_{a_j}([\gamma]\phi))$
\end{tabular}
\end{center}
\begin{center}
\begin{tabular}{l}
$M,w\models purpose(r,\gamma,c,\phi)$ \mbox{iff} \\ 
$\forall \alpha_i\exists a_j: M,w\models play(a_j,r,c) \wedge$\\ $CB_{A_c} (Salient(A_c:\gamma,c) \wedge DO(a_j:\alpha_i)\rightarrow Goal_{a_j}(\phi) \wedge B_{a_j}([\gamma]\phi)$
\end{tabular}
\end{center}
{\rm The purpose of an abstract sequence of actions} $\delta$=$\cup_{i=1...n}\beta_i$:\\
\begin{center}
\begin{tabular}{l}
$M,w\models purpose(\delta,c,\phi)$ \mbox{iff} $\forall \beta_i M,w\models purpose(\beta_i,c,\phi)$
\end{tabular}
\end{center}
\end{definition}

Using all of the definitions above we can now give the definition of the purpose of a social practice.
\begin{definition}{{\bf (Purpose of a Social Practice)}}\\
{\rm Let $c\in SP$ and let $\Delta=\{\delta: SC_c \rightarrow [\delta]EC_c \}$ then}\\
\begin{center}
\begin{tabular}{l}
$M,w\models purpose(c,\phi)$ \mbox{iff} \\
$\forall \delta\in\Delta: M,w\models purpose(\delta,c,\phi)$
\end{tabular}
\end{center}
\end{definition}
In other words the purpose of a social practice is $\phi$ if for all the action sequences from the start to an end of the social practice the purpose of that sequence in the context of the social practice is $\phi$. 

Using the above definitions we can now finally give a definition of plan patterns in a social practice $sp$.
\begin{definition}{{\bf (Plan Patterns Language)}}\\
{\rm A Plan Pattern of a social practice is an element of the set PP, which is the smallest set closed under}
\begin{gather*}
\gamma\phi \in PP\\
\gamma_1\phi_1,\gamma_2\phi_2 \in PP \Rightarrow \gamma_1\phi_1 +\gamma_2\phi_2 \in PP \\
\gamma_1\phi_1,\gamma_2\phi_2 \in PP \Rightarrow \gamma_1\phi_1 \&\gamma_2\phi_2 \in PP \\
\gamma_1\phi_1,\gamma_2\phi_2 \in PP \Rightarrow \gamma_1\phi_1 ;\gamma_2\phi_2 \in PP\\
%\gamma\phi \in PP \Rightarrow (\gamma\phi)* \in PP
\end{gather*}
\end{definition}
By defining the set of PP separately from abstract actions, and enforcing them to be of the specific format $\gamma\phi$ we can enforce that all plan patterns have the format of an abstract action $\gamma$ that achieves a particular goal $\phi$. This means that every part of a plan pattern has its own "objective" or landmark that it aims to achieve. This objective is not a coindicidental formula that happens to be true after performing the action, but is the reason for performing the action.
Let $\gamma\phi,\gamma_1\phi_1 \in PP$ then we use $\gamma_1\phi_1 \in \gamma\phi$ if $\gamma_1\phi_1$ occurs in $\gamma\phi$. We use $st(\gamma\phi)$ to refer to the first (complex) pattern of $\gamma\phi$. So, if $\gamma\phi$=$\gamma_1\phi_1;\gamma_2\phi_2$ then $st(\gamma\phi)=\gamma_1\phi_1$. If $\gamma\phi$ is constructed using one of the other constructs then $st(\gamma\phi)=\gamma\phi$.\\
This leads us to the following definition of a plan pattern of a social practice.
\begin{definition}{{\bf (Plan Patterns of Social Practices)}}\\
{\rm Let $sp\in SP$ and $A',A'' \subset A_{sp}$ and let $\Delta_{sp}=\{\delta: SC_{sp} \rightarrow [\delta]EC_{sp} \}$}\\
\begin{center}
\begin{tabular}{ll}
$M,w\models$& $planpattern(\gamma\phi,sp)$ \mbox{iff}\\
&$\gamma\phi\in\Delta_{sp} \ \mathrm{and}$\\
&$\ purpose(sp,\phi) \ \mathrm{and}$ \\
&$\forall\gamma_i\phi_i \in \gamma\phi: purpose(\gamma_i\phi_i,sp,\phi_i) \ \mathrm{and}$ \\
&$SC(sp,\phi) \rightarrow strategy(\phi,DO(A_{sp}:st(\gamma\phi),sp) \ \mathrm{and}$\\
&$\forall\gamma_1\phi_1;\gamma_2\phi_2 \in \gamma\phi:$\\
& $strategy(DONE(A':\gamma_1\phi_1),DO(A'':\gamma_2\phi_2),sp)$
\end{tabular} 
\end{center}
\end{definition}
This states that the overall plan pattern must have the same purpose as the social practice and that any sub patterns of the main plan pattern of a social practice are abstract (sequences of) actions for which the purpose in the context of the social practice is to reach the formula $\phi$ associated with that abstract action. Moreover, the plan pattern starts automatically when the start condition of the social practice is true and all actors involved have a common belief that the social practice is active. And if the plan pattern specifies two abstract actions in a sequence there is an expectation that when the first part is done (by a subset $B$ of the agents) the second abstract action will be done. (See further down in this section for more explanation on the strategy relation).
The following figure shows a graphical representation of a plan pattern for taking the children to school.
\begin{figure}
\centering
\includegraphics[width=0.9\columnwidth]{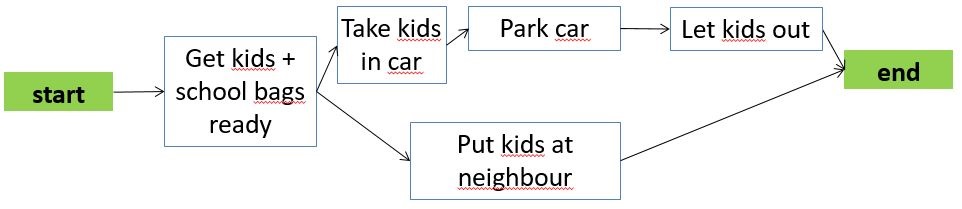}
\caption{Plan Pattern for taking kids to school}
\label{fig:pp}
\end{figure}
This plan pattern can be describe formally by first describing the pattern as a combination of the basic complex actions that constitute this pattern. 
\[ \gamma\phi = \gamma_1\phi_1;((\gamma_2\phi_2;\gamma_3\phi_3;\gamma_4\phi_4)+\gamma_5\phi_5)\]
where:\\
\begin{center}
\begin{tabular}{l}
$\gamma_1$ = get-kids \& get-bags and $\phi_1$ = kids-with-bags-ready\\
$\gamma_2$ = everyone-gets-in-car;drive-car and $\phi_2$ = car-at-school\\
$\gamma_3$ = park-car and $\phi_3$ = car-at-safe-place\\
$\gamma_4$ = kids-out-car;kids-go-school and $\phi_4$ = kids-at-school\\
$\gamma_5$ = kids-go-neighbour;neighbour-drives-to-school and $\phi_4$ = kids-at-school\\
$\phi$ = kids-are-safe $\wedge$ kids-at-school $\wedge$ time before 9:00
\end{tabular}
\end{center}
Notice that each of the parts itself usually is again a complex action involving many actions of different actors. However, within these parts (or phases) we do not have to have a specific purpose for each action anymore and thus there is more freedom on how to fill it in. Notice as well that there is no prescribed level of detail. The more detailed the plan pattern parts are made, the more fixed becomes the interaction. The bigger the phases are the more freedom for the actors to fill in these phases as they like (fits their own goals). 

Many elements of the social practice can be seen as expressing a kind of expectation. As stated before expectations with social practices come in different forms. The plan patterns can be seen as expectations because they indicate the general patterns of behavior that are expected. However, we also have more specific expectations. Although not all parts of a social practice are fixed there are some points where specific types of actions are expected. E.g. if in a greeting practice one person extends his hand it is expected that the other person shakes the hand.
We use the ADICO grammar proposed by Ostrom to specify rules, norms, and strategies \cite{crawford1995grammar}. ADICO statements
are formed using the following five components: Attribute (or Acting entity), Deontic, aIm (or Intention), Condition , and Or else (or sanction). In ADICO Strategies are statements including only the acting entity, intention, and condition (AIC); Norms
include the acting entity, deontic, intention, and condition (ADIC); and Rules include all 5 components (ADICO). The norms and startegies as described through the ADICO grammar can nicely be formalized using the logic that we already presented. We start with defining strategies.
\begin{definition}{{\bf (Strategies)}}\\
{\rm Let $\gamma \in ActExp$ and let $B\subset A_{sp}$ be a set of actors (possibly one)}\\
\begin{center}
\begin{tabular}{l} 
$M,w\models strategy(\phi, DO(B:\gamma),sp)$ \mbox{iff}\\ 
$M,w\models CB_{A_{sp}}(CB_B(\phi)) \wedge CB_{A_{sp}}(active(sp)) \rightarrow CB_{A_{sp}}(DO(B:\gamma))$
\end{tabular}
\end{center}
\end{definition}
Thus if all actors (involved in the social practice $sp$) believe that the social practice is active and $B$ believes the condition $\phi$ then they all believe that all actors in $B$ will perform their part of $\gamma$ next. When the condition $\phi$ is of the form $DO(C:\gamma_1)$ or $DONE(C:\gamma_1)$ the strategy indicates a synchronization or sequencing of actions repsectively. If $\phi$ is representing some state to be true it is expected that $B$ takes the initiative to perform some action when a certain state is reached.\\
Note that the above definition entails that if actor $a \in B$ then she believes that if she does her own part of $\gamma$ the rest of $B$ will do their part of $\gamma$.\\
Instead of the $DO(B:\gamma)$ it is also possible to use the weaker $H_B(DONE(B:\gamma))$ which means that $B$ is attempting to perform $\gamma$, but the outcome is not guaranteed. 
The strategies in a social practice are mostly used to define small parts of the interaction that always are done in the same way and thus have a kind of fixed protocol. E.g if a child in class raises her hand then the teacher will respond and ask what she wants. This can happen at any time and thus is not bound by the plan pattern, but it can fill in part of the plan pattern, of course. As we have seen in the definition of plan patterns, the strategies are also used to make sure the social practice moves from one phase of the plan pattern to the next (without having to explicitly coordinate this). E.g. when the children and the parent are in the car, they start driving to school.

As stated before norms are also a kind of expectations. However, instead of a belief that an action will be taken they lead to a normative statement that an action \emph{should} be taken. Based on the dynamic logic we can define obligations, prohibitions and permissions very easily as follows.
\begin{definition}{{\bf (Deontic operators)}}\\
\begin{gather*}
    O(A:\gamma) = [\overline{A:\gamma}]V\\
    F(A:\gamma) = [A:\gamma]V\\
    P(A:\gamma) = \neg[A:\gamma]V
\end{gather*}
\end{definition}

In the context of social practices, the norms are connected to the roles of the social practice. Thus any actor that enacts a role in the social practice should follow the norms specified for that role. So, the definition of norms following the ADICO framework is as follows.
\begin{definition}{{\bf (Norms)}}\\
\begin{center}
\begin{tabular}{l}
$M,w\models O(r,\phi,\gamma,\rho)$ \mbox{iff}\\
$M,w\models \forall a: play(a,r) \wedge B_a(\phi) \rightarrow (O(a:\gamma) \wedge [\overline{a:\gamma}]O(a:\rho)$\\
$M,w\models F(r,\phi,\gamma,\rho)$ \mbox{iff}\\
$M,w\models \forall a: play(a,r) \wedge B_a(\phi) \rightarrow F(a:\gamma) \wedge [a:\gamma]O(a:\rho)$\\
$M,w\models P(r,\phi,\gamma,\rho)$ \ \mbox{iff} \ $M,w\models\forall a: play(a,r) \wedge B_a(\phi) \rightarrow P(a:\gamma)$
\end{tabular}
\end{center}
\end{definition}
Looking at the definition of the norms more closely reveals that they are still somewhat simple. The obligations have to be fulfilled the moment they become active. In normal life when an obligation becomes active one has some time to react and fulfill it. In early work \cite{dignum1996} we have already shown how to model some temporal aspects into the norms while using the dynamic logic as underlying formalism. Of course, it is easier to express temporal properties, like deadlines, using temporal logic as basis \cite{DignumKuiper1997,dignumdignum2004}. We leave the current version as a placeholder for more complex formalizations that could be inserted when necessary.

Besides the above elements that are used in several parts of the formal representation of social practices we also have some more ontological elements that are used to describe meanings and resources. First of all we extend the language $\mathcal{L}$ with the counts-as relation $countsas(c,r:\alpha_1,\alpha_2)$
to mean that if any actor is performing action $\alpha_1$ in context $c$ in the role of $r$ this is seen as performing action $\alpha_2$. E.g. the lecturer raising her hand counts as starting the lecture. The semantics of this construct is given as a kind of "is-a" relation within context. This is a kind of simple form of the formal definition of this concept as given in \cite{Grossiphd}. In this work also relations between contexts are taken into account as well as combinations of concepts. Within the context of social practices the "is-a" feature that connects the physical actions with social actions and their effects is the only thing we need. We still use the "counts-as" term in order to keep further connections with the more elaborate work as from \cite{Grossiphd} possible for future work.
\begin{definition}{{\bf (Counts-as)}}\\
\begin{center}
\begin{tabular}{ll}
$M,w\models$ & $countsas(c,r:\alpha_1,\alpha_2)$ \mbox{iff} \\
     & $\forall w'\in \llbracket c \rrbracket: \forall a: rea(a,r,w')$ \\
     & $M,w'\models [\alpha_2]\phi$ then $M,w'\models [a:\alpha_1]\phi \wedge CB_{A_c}([a:\alpha_1]\phi)$
\end{tabular}
\end{center}
\end{definition}
Given this definition we can substitute $\alpha_1$ for $\alpha_2$ wherever it is needed. E.g. if there is an obligation on the lecturer to start the class, then Ann (the lecturer) can fulfill the obligation by raising her hand. The definition does not state that the two actions are equal! It can be (and usually is the case) that $\alpha_1$ has more effects than $\alpha_2$. E.g. by raising her hand Ann starts the class, but now also has her hand up in the air (which is not an effect of starting the class). The last part of the definition of the counts-as ensures that it also is common belief that $\alpha_1$ performed by someone playing role $r$ in the context $c$ has $\phi$ as effect. This is especially important if the effect is a non-tangible or social effect which is not directly observable. This part makes sure that all actors involved have a common belief on the social effect of actions in a context.

The second concept that we need is the fact that an action is promoting or demoting a certain value. E.g. students talking during a lecture demotes the value of respect. This concept is used to indicate some abstract social effects of actions. Again this relation is context dependent. We extend the language $\mathcal{L}$ with a set of special constants $Value=\{v_1,...v_n\}$ that denote values and $promotes(sp,r:\alpha,v_i)$ and $demotes(sp,r:\alpha,v_i)$. 
It indicates that if an actor plays the role $r$ and performs the action $\alpha$ she promotes or demotes the value $v$. Again, we will give a very simple semantics for this construct based on a more elaborate characterization of the promotes relation given in \cite{Weidephd,dignum2019}.
\begin{definition}{{\bf (Promotes and Demotes Value)}}\\
{\rm We extend the model $M$ with a set $Val=\{<_1,...,<_n\}$ of orderings where each $<_i$ is a partial order on $\mathbb{W}$ indicating the preference of the states according to that value.}
\begin{center}
\begin{tabular}{ll}
$M,w\models$&$promotes(sp, r:\alpha,v_i) \ \mbox{iff} $\\
&$w\in \llbracket sp\rrbracket \ \mbox{and} \ \forall a: rea(a,r,w) \ \mbox{then} \ \forall w'\in \llbracket a:\alpha\rrbracket_R <_i(w,w')$\\
$M,w\models$&$demotes(sp, r:\alpha,v_i) \ \mbox{iff} $\\ 
&$w\in \llbracket sp\rrbracket \ \mbox{and} \ \forall a: rea(a,r,w) \ \mbox{then} \ \forall w'\in \llbracket a:\alpha\rrbracket_R <_i(w',w)$
\end{tabular}
\end{center}
\end{definition}
The definition states that an action $\alpha$ performed by an actor playing role $r$ promotes (resp. demotes) a value $v_i$ if the action $\alpha$ leads from state $w$ to a state $w'$ that is better (resp. worse) according to value $v_i$.\\
So values indicate a kind of general preference on the worlds according to the perspective of that value. The promotes predicate indicates the "value" of actions on the social level and are used to connect the social practice to a wider social context in which certain type of behavior is valued and other behavior is to be avoided. E.g. we are environment conscious and promote sustainable behavior. A social practice taking the children to school by bike will promote this value, while taking them by car demotes this value. However, using the car can promote safety , which also might be important.

Finally we also introduce the concept of affordances here. Affordances are used to describe the type of actions that are normally expected to be performed with an object and can also be used to describe types of objects in an abstract way. E.g. an object that affords to sit (which can be a chair or a bank or ...). Again we extend the language $\mathcal{L}$ with a set of special constants $Obj=\{o_1,...,o_m\}$ denoting the objects and $affords(O,\alpha,c)$ which is the affordance relation stating that the set of objects $O\subset Obj$ affords action $\alpha$ in context $c$. Here we take it that all objects of the set $O$ together afford the action. The set can consist of zero, one or  more elements. If the set is empty, the action does not require any object to be performed.
In order to indicate that objects are in the place where the social practice takes place we use the predicate $available(O,c)$. We will not give semantics for these predicates as they should be tied to a theory on geometrical and physical properties of objects, which is not the focus of this paper. Having all these parts in place we can tie affordances, objects and capabilities to the possibility of an action being performed in a certain situation. 
\begin{definition}{{\bf (Affordances)}}\\
{\rm Let $O=\{o_1,...,o_m\}$ and $O\subset O_c$ then}
\[M,w\models affords(O,\alpha,c)\wedge available(O,c)\wedge play(a,r,c) \rightarrow G_a\alpha\]
\end{definition}
This states that if the set of objects $O$ that affords an action $\alpha$ is available and the actor $a$ is playing the role $r$ in the same context then the action can be performed by the agent playing the role. It does not guarantee any postcondition of the action to hold. But for any social practice we can specify the following expectation:
\[M,w\models CB_{A_c}(affords(O,\alpha,c)\wedge available(O,c)\wedge play(a,r,c)) \rightarrow CB_{A_c}(G_a\phi)\]
which states that there is a common belief that if all the preconditions hold then it is common belief that the action will lead to some predetermined effects $\phi$. This specifies in a very cautious way the postconditions of actions, given the preconditions are fulfilled.

\section{Representing and Using Social Practices}
\label{applications}
\begin{figure}[ht]
    \centering
    \includegraphics[width=\textwidth]{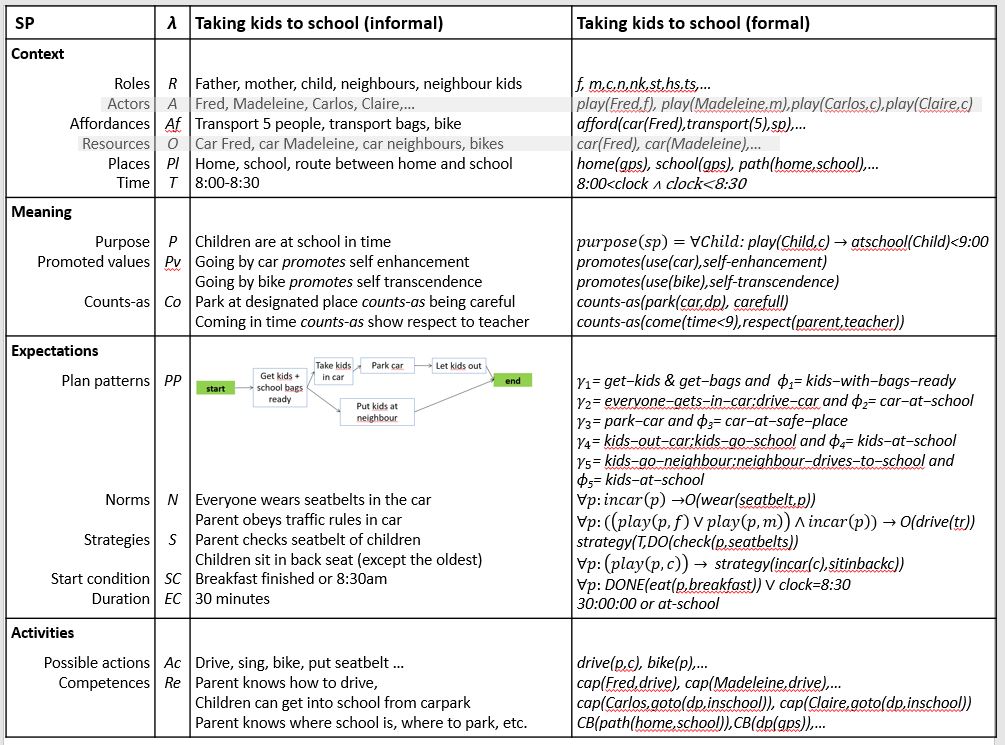}
    \caption{Social Practice of taking kids to school}
    \label{fig:SPform}
\end{figure}

In the previous sections we have described all ingredients for representing instances of social practices, which allows us to now formalize all elements of table \ref{fig:socialpractice} completely. This leads to table \ref{fig:SPform}.
Formally an instance of a social practice can now be defined as as a tuple of all the elements from the table. 

We now come back to the fact that this actually describes an instance of a social practice. The fact that we have defined the logic based on a kripke framework fitting dynamic logic, means that we always talk about concrete actions being performed and sequences of actions are between concrete situations that are linked to a specific world. Using our example, we talk about a specific instance of taking the children to school, which is on a specific day, involves specific people and a very specific school. What we now want to do is to generalize over all these instances of a social practice to get a more general (usable) social practice of taking children to school.
We do this by defining a social practice in terms of its instances.
\begin{definition}{{\bf (Social Practice)}}\\
{\rm Let} 
\begin{center}
\begin{tabular}{ll}
$sp_i=$&$(R_{sp_i},A_{sp_i},O_{sp_i},Af_{sp_i},Pl_{sp_i},P_{sp_i},Pv_{sp_i},Co_{sp_i},PP_{sp_i},$\\
&$N_{sp_i},S_{sp_i},SC_{sp_i},EC_{sp_i},Ac_{sp_i},Re_{sp_i})$
\end{tabular}
\end{center}
{\rm then we say sp is a social practice based on $sp_i$ written as
$SocialPractice(sp,sp_1,...,sp_n)$ iff}\\
\begin{enumerate}
\item $\forall i,j=1...n: R_{sp_i}=R_{sp_j}=R_{sp} \ {\rm and}$\\
$\exists a_i\in A_{sp_i},r\in R_{sp_i}: play(a_i,r,sp_i) \ {\rm then} \ \exists a_j\in A_{sp_j}: play(a_j,r,sp_j)$
\item $\forall i,j=1...n: affords(O_k,\alpha_l,sp_i) \ {\rm then}$ \\
$\exists O_q: \ affords(O_q,\alpha_l,sp_j) \ {\rm and} \ \exists O: \ affords(O,\alpha_l,sp)$
\item $P_{sp}=\bigcup_i=1...n P_{sp_i} \ {\rm and} \ \bigcap_i=1...n P_{sp_i}\not=\emptyset$
\item $Pv_{sp}=\bigcup_i=1...n Pv_{sp_i} \ {\rm and} \ \bigcap_i=1...n P_{sp_i}\not=\emptyset$
\item $Co_{sp}=\bigcup_i=1...n Co_{sp_i} \ {\rm and} \ \bigcap_i=1...n P_{sp_i}\not=\emptyset$
\item $\forall i=1...n: PP_{sp_i}=PP_{sp}$
\item $\forall i=1...n: N_{sp_i}=N_{sp}$
\item $\forall i=1...n: S_{sp_i}=S_{sp}$
\item $SC_{sp}=\bigvee_i=1...n SC_{sp_i}$
\item $EC_{sp}=\bigvee_i=1...n EC_{sp_i}$
\item $Ac_{sp}=\bigcup_i=1...n Ac_{sp_i}$
\item $\forall i=1...n: Re_{sp_i}=Re_{sp}$
\end{enumerate}
\end{definition}
The definition states that the social practice instances should be the same in many respects. For the context we only require that in all practice instances we have the same roles and that if there are actors playing a role there should be actor playing that same role in all other practice instances. This basically states that if a role is relevant in one instance (because an actor plays it) it should be relevant in all other instances. So, we cannot fulfill the requirement of having the set of roles being the same in all instances by just taking all possible roles for all instances. The affordances requirement makes sure that, even if different objects are present in different instances, the same set of actions is afforded by the objects in all instances.\\
The elements from the meaning block can differ, but need to have a common core. The reason not to have them all equal is that in specific contexts there can be additional values promoted or actions used for social effects. \\
The plan patterns, norms and strategies should be the same for all instances. They are already abstracting away from specific actions. So, different instances might have different ways of following the plan pattern, but they all follow the same pattern.\\
The start conditions and end conditions were made very specific to fit each instance. Therefore they have to be generalized (by taking the disjunctions of all of them) to generate the start and end conditions of the general social practice. The same holds for the actions that can be expected. these can be very specific for different instances. The disjunction allows to generalize over the specifics of each instance.\\
Finally, the requirements on competences should be such that they are the same for all instances. These are assumptions that are used by the actors participating in the social practice and should thus not differ between instances.

\subsection{Properties and Use of Social Practices}
\begin{figure}
    \centering
    \includegraphics[width=0.7\textwidth]{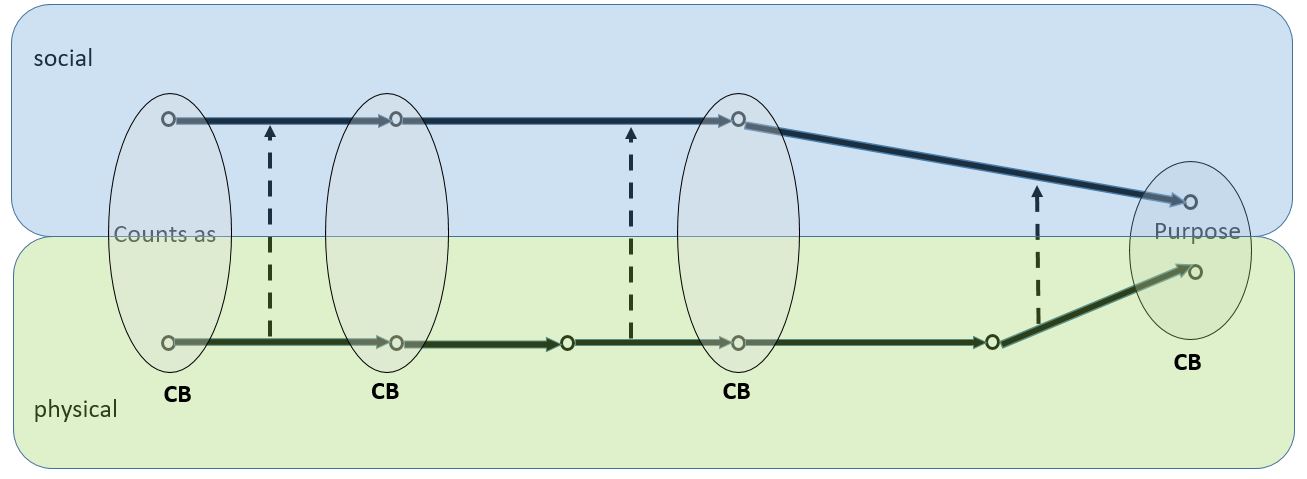}
    \caption{parallel social and physical effects}
    \label{fig:socialreal}
\end{figure}

Given the large amount of definitions on social practice elements we can now also define a number of properties of social practices. We start with three properties of the social practices that are not necessarily guaranteed by the formalism, but are required for the social practice to be useful. In the following we use $\Delta_{sp}=\{\delta: \delta\in PP_{sp} \wedge SC_{sp} \rightarrow [\delta]EC_{sp} \}$.
\begin{definition}{{\bf (Feasible Social Practice)}}\\
 A social practice $sp$ is feasible ($\emph{feasible}(sp)$) iff:
\[\exists \gamma\phi\in\Delta_{sp} \ {\rm and} \forall\alpha_i\in \gamma\phi: \exists a\in A_{sp}: G_a\alpha_i
\]
\end{definition}
This states that there is at least one sequence of actions from start to end conditions such that it follows the plan pattern of the social practice and every action can be performed by some actor (notice we don't state that $cap(a,\alpha_i)$ but the stronger $G_a(\alpha_i)$). This means that there is a way to execute the social practice. This is the minimum that we would like. But it would also be nice if one could execute the social practice without violating any of the norms connected to the social practice.
\begin{definition}{{\bf (Normative SP)}}\\
A social practice $sp$ is normative ($\emph{normative}(sp)$) iff:
\begin{center}
\begin{tabular}{l}
$feasible(sp) \wedge$\\
$\exists \gamma\phi\in\Delta_{sp} \ {\rm and} \forall\alpha_i\in \gamma\phi: (\neg (F(r,\phi_i,\alpha_i,\rho_i) \wedge \phi_i \wedge DO(r:\alpha_i)) \wedge$\\
$O(r,\phi_i,\alpha_i,\rho_i) \wedge \phi_i\rightarrow DO(r:\alpha_i)$
\end{tabular}
\end{center}
\end{definition}
This just states that there is a way to execute the social practice without violating any of the norms of the social practice whenever the norms are active. Notice that actions might be obliged or prohibited during the social practice, but have no influence if their activation condition is not true at that moment.\\
The next definition states that we have completely defined the relation between the physical and social parts of the actions. In order to define this we have to split our states into a social and physical part $w=<w_p,w_s>$ (or brute facts and social facts as Searle calls them  \cite{searle1995construction}).We call a formula $\phi$ social ($social(\phi)$) iff 
\[ M,<w_p,w_s> \models \phi \equiv M,w_s\models \phi \]
The social formulas decribe the social world, which is only indirect observable. This means that we only know the formula is true because we have seen action that has this formula as social effect. Completeness of a social practice means that all social formulas that are part of a purpose in the plan pattern of the social practice are either a direct effect of an action or a (social) effect of an action connected to an action with a count-as relation.
\begin{definition}{{\bf (Complete SP)}}\\
A social practice $sp$ is complete ($\emph{complete}(sp)$) iff:
\begin{center}
\begin{tabular}{l}
$feasible(sp) \wedge$\\
If $\exists \gamma\phi\in\Delta_{sp} \ {\rm and} \forall\phi_i\in \gamma\phi: social(\phi_i) \ {\rm then}$\\
$\exists \gamma_i\in \gamma\phi: [\gamma_i]\phi_i \ {\rm or} \
\exists \alpha_i\in \gamma\phi: countsas(sp,r:\alpha_i,\alpha_j,c) \ {\rm and} [\alpha_j]\phi_i$
\end{tabular}
\end{center}
\end{definition}
If a social practice is complete then all the participants in the social practice have a common belief of both the physical as well as social state during the execution of the social practice. The physical state is guaranteed through the common belief on the plan pattern, while the social state can be derived from the combination of actions and the social effects of the social actions connected to them through the counts-as relation as shown in figure \ref{fig:socialreal}

In the remainder of this section, we will briefly describe how an actor can use social practices to deliberate about its context and plan its actions. It should be emphasized that:
\begin{enumerate}
    \item A social practice is not a very complex but complete specification of an interaction. It is a rich source of information of elements that can be used by actors to plan their interactions with some guaranteed outcomes if they follow the plan patterns, the strategies and the norms.
    \item We make no assumptions about the actors using the social practices. In practice we assume they are actors that use goals and plans to deliberate about their (inter)actions. However, this does not have to be true. Thus the specification is not just an abstraction of an interaction protocol, but rather a specification of interactions with some holes (to be filled by the agents by choosing the fitting specific actions). 
\end{enumerate}
In \cite{miller2017planning} we have shown how a standard Multi-Agent epistemic planner can be used to execute social practices and the effects the use of the social practice has on the efficiency of the planning and the robustness of the plan. Here we just walk through the simple social practice example of taking the children to school given in \ref{fig:socialpractice}.

Let's take the perspective of Fred (the father of the children). Assuming this actor to follow a BDI-like architecture, the goal ``get the children to school'' would be part of the Intentions of Fred. We further assume that both the social practice described in table \ref{fig:socialpractice} and the prerequisite knowledge are part of Fred's beliefs (and of his children). Fred also has the capability to drive a car leading Fred to belief that he could fulfill the role of \textit{driver}. By using a social practice-based deliberation process, once Fred beliefs to be himself in the context of the social practice of taking the children to school (i.e. if it is 8:30 or breakfast is finished), he can form a plan for taking the children to school, based on the expectations, meaning and activities the social practice describes for actors of role `parent'. This plan is very simple, Fred goes to the car, let's the children in the car, drives to school, parks in the designated parking and lets the children out. While Fred is driving he can interact with the children using strategies and norms. He can choose to enquire about their plans for the day at school or talk about a soccer match or answer questions of the children. Based on the plan pattern of the social practice Fred can conclude that by fulfilling its part, the desired result will be achieved. Very important in this respect is that he can assume the children will e.g. take their bags. Thus Fred does not have to plan for all possible actions or reactions of the children, but can assume a standard behaviour. However, the social practice also indicates that after the first part, the children and the bags are ready. Thus Fred can check if the bags are there as well. Thus the social practice gives focus of attention and gives times at which to check for specific states to hold. The only thing he still needs to plan is the exact route from house to school, which might be standard and directly available or needs to be planned based on some construction work, expected traffic, etc. Now, suppose that Fred does his part and gets in the car, but the car does not start. The social practice now gives some handles on what can be done as well. First of all, the car is an object that affords driving the kids to school. Is there another object with the same affordance? E.g. the car from Madeleine (Fred's partner) might also be available. Thus Fred can simply replan and take the car of Madeleine.\\
Social effects are not visible but can be assumed given the social practice. E.g. leaving in time will result in being in time at school and thus showing respect to the teacher. Thus following the social practice successfully also guarantees some social effects, even though these might not be observed directly.

\section{Discussion and Conclusions}
\label{conclusion}
In this paper we gave a formalization of social practices that aims to support actors planning \emph{social} interactions. We have briefly shown that \emph{feasible} social practices can simplify the amount of actions that have to be planned by an actor and that it can make assumptions about interactions with other actors in the social practice without explicitly having to coordinate. 

From the paper it has become obvious that social practices are quite complex. They contain many elements that are also interwoven. Would it not be simpler to just use a collection of protocols to regulate the interaction? In some applications this might certainly be true. For interactions between agents in Multi-Agent Systems that are designed for a specific purpose and operate in a stable environment one does not need a sophisticated notion such as social practices. Note that a social practice can collapse into a fixed protocol when the plan pattern describes all specific actions and their intended effects. So, a social practice can be designed to give more or less flexibility.\\
Social practices become interesting when software agents have to interact in dynamic environments where standard interactions do not always work or when agents interact with humans. The social practices have a level of abstraction from the actual actions to be taken that gives freedom to react to the current situation without crashing the interaction protocol. Due to all the additional information about the context, alternatives can be found, while the expectations give guidance to which alternatives are preferable. Finally, when interacting with humans it pays off to follow the social practice that the human is used to, because humans use social practices most of the time. Thus one can easily make many assumptions on how the human will behave without having to deliberate about them explicitly.\\
Another important aspect is the automatic social effects of a social practice. It might be possible to fulfill the purpose of a social practice in different ways as well, but following the social practice guarantees certain social effects that the actor does not have to deliberate or plan separately. This becomes very important when interacting with humans, especially if the interaction is a long term one.\\
We have shown in some very preliminary work how all these aspects can actually be used in implemented systems. In \cite{SALVE2016} we have shown how social practices can be used to structure natural dialogues. In \cite{Robophil2018} we have shown how social practices can add value to interactions for social robots. 

As we already indicated in the introduction social practices are social constructs that emerge from repeated interactions between people. In some way this feature is replicated in our formalism through the fact that a social practice is associated to the tuple of all its instances. However, where we had to indicate all kinds of constraints on how a social practice relates to its instances, in social practices in human society these constraints are fulfilled by nature of how the social practices themselves emerge from the repeated interactions. The same holds for the properties of social practices. Because we constructed the social practice structure we have to impose a property like "feasibility". In human practice the social practice emerges from the instances. Each instance is by definition feasible and thus the abstract social practice by nature is feasible. This type of property is not easily represented using a formal logic. Similarly, the way a social practice is performed will not be a random choice between all allowed action sequences. The more successfull ones will be repeated most. This would lead to a more probabilistic representation of the structure based on experiences. Again this aspect has not yet been explored but could be done by extending the logic with a probabilistic component. However, one might also take the formalized social practice structure as a given basic notion and use a probabilistic formalism to model this aspect of adaptation to the circumstances separately.

An reservation against using social practices for AI systems is that they are very complex and therefore are hard to obtain for real systems. Although we did not specify a methodology for obtaining social practices for AI systems yet, in our experience it is actually relatively simple to sollicit them from the intended users of the system. People are used to social practices and can easily (and precisely) describe their components when asked for them. In top of that one can also observe them when they are enacted and easily infer parts from the paper trails when they are performed in formal organizations. Once there are a number of generic social practices available the AI systems can also learn new social practices based on existing ones. Some preliminary work on this has been set up, but a generally usable tool is part of future work.

Another important area for future work is the development of learning mechanisms that enable efficient matching of sensing information and beliefs to the components of a social practice description. E.g. when we enter a shop we assume there will be a sales person around. Thus we start looking for a person that might fulfill that role. If we give a presentation in a room we start looking for a projector and a cable to connect the laptop. In this way, the fact that we assume we are in a social practice can drive our sensing and perception to find objects and persons or facts that fill in the various elements of that social practice. Especially for social robots that act in a physical, dynamic and open environment this focus of attention supports their coping with complex situations.

We finish this paper by restating that the main goal was to provide an unambiguous and precise account of the concept of social practices for the purpose of using it in social applications of AI systems. We can conclude that we have now a solid basis for the implementation of social practices in many different application areas.

\bibliographystyle{plain}
\bibliography{social}

\end{document}